# Lyra: An Efficient and Expressive Subquadratic Architecture for Modeling Biological Sequences

Krithik Ramesh*[1,2], Sameed M. Siddiqui*[1,3], Albert Gu[4], Michael D. Mitzenmacher[5]†, Pardis C. Sabeti[1,6,7,8]†

[1] Broad Institute of MIT and Harvard, Cambridge MA, 02142

[2] Massachusetts Institute of Technology, Cambridge MA, 02139

[3] Computational and Systems Biology Program, Massachusetts Institute of Technology, Cambridge MA, 02139

[4] Machine Learning Department, Carnegie Mellon University, Pittsburg, PA, 15213

[5] School of Engineering and Applied Sciences, Harvard University, Cambridge, MA, 02138

[6] Department of Organismic and Evolutionary Biology, Harvard University, Cambridge, MA, 02138

[7] Department of Immunology and Infectious Diseases, Harvard T.H. Chan School of Public Health, Harvard University, Boston, MA, USA, 02115

[8] Howard Hughes Medical Institute, Chevy Chase, MD, USA, 20815

correspondence: {krithik, sameed}@mit.edu, michaelm@eecs.harvard.edu, pardis@broadinstitute.org

* these authors contributed equally

† these authors jointly supervised this work

## Abstract

Deep learning architectures such as convolutional neural networks and Transformers have revolutionized biological sequence modeling, with recent advances driven by scaling up foundation and task-specific models. The computational resources and large datasets required, however, limit their applicability in biological contexts. We introduce Lyra, a subquadratic architecture for sequence modeling, grounded in the biological framework of epistasis for understanding sequence-to-function relationships. Mathematically, we demonstrate that state space models efficiently capture global epistatic interactions and combine them with projected gated convolutions for modeling local relationships. We demonstrate that Lyra is performant across over 100 wide-ranging biological tasks, achieving state-of-the-art (SOTA) performance in many key areas, including protein fitness landscape prediction, biophysical property prediction (e.g. disordered protein region functions) peptide engineering applications (e.g. antibody binding, cell-penetrating peptide prediction), RNA structure analysis, RNA function prediction, and CRISPR guide design. It achieves this with orders-of-magnitude improvements in inference speed and reduction in parameters (up to 120,000-fold in our tests) compared to recent biology foundation models. Using Lyra, we were able to train and run every task in this study on two or fewer GPUs in under two hours, democratizing access to biological sequence modeling at SOTA performance, with potential applications to many fields.

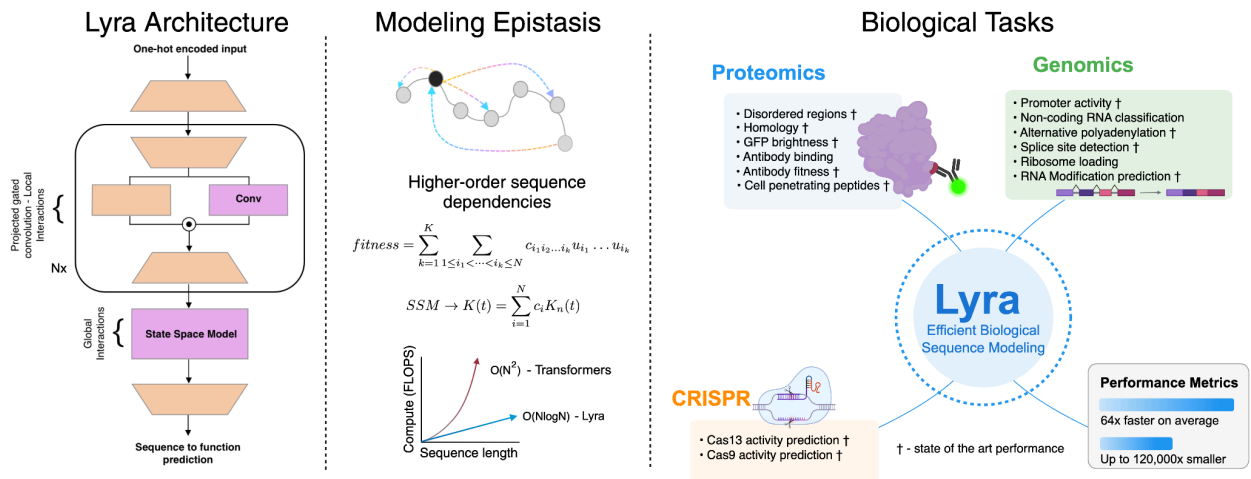

**Graphical abstract.** (Left) The Lyra architecture introduces an efficient approach for biological sequence modeling, combining projected gated convolutions for local feature extraction with state space models (SSMs) for capturing long-range dependencies. (Center) Lyra addresses the fundamental challenge of modeling epistasis—complex interactions between sequence elements—by leveraging SSMs' intrinsic ability to approximate polynomials. This mathematical alignment enables efficient $O(N \log N)$ scaling with sequence length, compared to the $O(N^2)$ complexity of attention-based approaches. (Right) We demonstrate Lyra's broad utility across biological domains: in proteomics, genomics, and CRISPR applications. Without pre-training and using orders of magnitude fewer parameters (up to 120,000-fold in our tests), Lyra matches or exceeds state-of-the-art performance across many tasks while providing substantial speedups compared to Transformer-based foundation models (on average 64.18x faster for batch size 2) in inference time.

# Introduction

The interpretation and modeling of biological sequences are central challenges in computational biology, with profound implications for understanding molecular function and evolution. At their core, biological sequences—whether DNA, RNA, or proteins—encode explicit instructions that determine molecular properties. Machine learning (ML) approaches seek to uncover these sequence-to-function relationships by modeling how primary sequence determines structural stability, fitness landscapes, and molecular activities[1–6]. Deep learning approaches have attempted to decode this inherent "grammar" by capturing both local patterns and long-range dependencies in biological data[6–10]. This insight allows us to predict how sequence variations affect biological function across scales, from protein folding to cellular regulation[1,11–14].

Deep learning models such as convolutional neural networks (CNNs) and Transformers have become powerful tools for biological sequence modeling, with each excelling in different domains[1,2,15–17]. CNNs excel at identifying local patterns and maintain efficient subquadratic $O(N*K)$ scaling with sequence length *N* and kernel size *K*[18–20]. Transformers excel at capturing long-range dependencies through self-attention mechanisms, enabling pairwise comparisons between distant residues, but require quadratic $O(N^2)$ scaling with sequence length[21–23]. Transformer-based models, such as AlphaFold2 [1], have demonstrated remarkable success in tasks like protein structure prediction by leveraging the evolutionary insight that sequence homology implies structural conservation[1,24]. However, Transformers often struggle with modeling local motifs, and their quadratic computational complexity limits their scalability. Hybrid architectures, such as Enformers[2,25], have been developed to combine CNNs for local context modeling with Transformers for global interactions, although they remain constrained by Transformer scaling limitations[26].

Achieving high performance in either Transformer-only or hybrid models frequently requires immense scale—often exceeding billions of parameters—as demonstrated by models like ESM3[27]. This reliance on scaling to capture task-specific patterns often falls short in biological systems due to a mismatch between the limited data available in many biological tasks and the scale required to learn the nuanced sequence-function relationships[9,28]. This highlights the need for continued innovations in model efficiency and scalability[29–31].

To address these challenges, we sought to identify intrinsic biological phenomena with well-defined mathematical structures that could provide a tractable foundation for modeling biological sequences. Epistasis—the influence of mutations on each other within a sequence—is one such phenomenon. [32–35]. While empirically complex and not fully understood, epistatic interactions can be characterized as combinations of individual and joint effects. This characterization allows us to represent these influences as the aggregated contributions of different sets of amino acids at specific positions, which corresponds to a class of multilinear polynomials [36]. This provides a principled approach for navigating the combinatorially vast space of sequence-to-function relationships.

Building on this polynomial lens, we identify state space models (SSMs) as a natural fit for sequence-to-function modeling[31,37–40]. We draw a mathematical connection between the hidden dimensions of SSMs and their capacity to model polynomials, highlighting a framework where hidden states effectively approximate the polynomial terms that govern epistatic relationships. This approach offers two key advantages over Transformers. First, modeling in the Fourier domain with Fast Fourier Transform (FFT) convolutions aligns closely with the mathematical structure of epistasis, providing a global representation of biological sequence interactions. Second, SSMs maintain this expressivity while scaling subquadratically with sequence length, offering greater computational efficiency compared to the quadratic complexity of self-attention. To complement the global dependencies captured by SSMs, we identify gated depthwise convolutions as an efficient mechanism for extracting local sequence features while enhancing expressivity through multiplicative gating. This approach allows for adaptive feature selection, ensuring that relevant sequence motifs are dynamically emphasized while maintaining computational efficiency

Here we develop an ML architecture, Lyra, which integrates gated convolutions with SSMs, creating a hybrid approach that efficiently captures both local and global relationships. This design achieves subquadratic computational complexity while maintaining the ability to model complex biological sequence-to-function relationships. Through careful analysis of the model's parameterization, we demonstrate how Lyra decomposes complex epistatic interactions into tractable components, providing insights into both the model's function and the underlying biological mechanisms it captures.

We evaluate Lyra through an extensive set of biological sequence modeling tasks that span multiple scales of complexity. At the protein level, we assess performance on fundamental biophysical properties (such as disordered regions), viral protein identification, and challenging protein engineering applications (including antibody binding, green fluorescent protein [GFP] fluorescence, and cell-penetrating peptides). At the nucleic acid level, we examine RNA function prediction (including splice sites, alternative polyadenylation, and ribosome loading), promoter activity prediction, and Clustered Regularly Interspaced Short Palindromic Repeats (CRISPR) genome editing efficiency (for both Cas9 and Cas13 systems). Beyond benchmarking performance, we perform detailed ablation studies to investigate the contributions of the architectural components in Lyra.

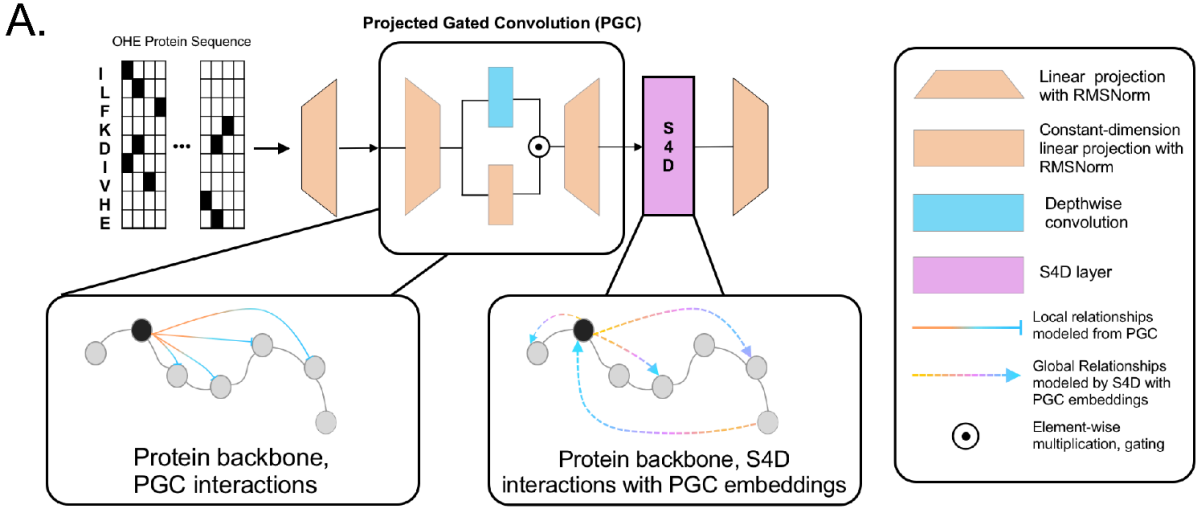

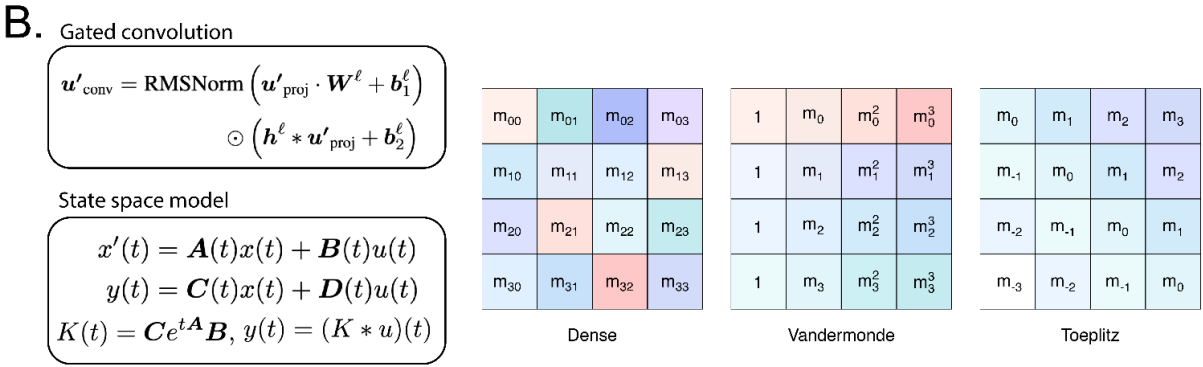

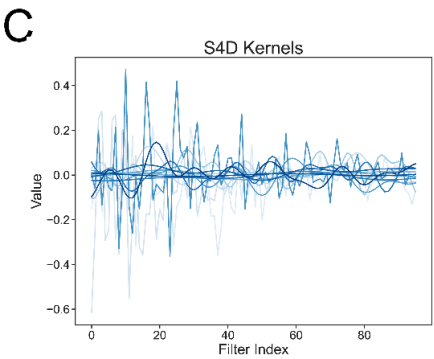
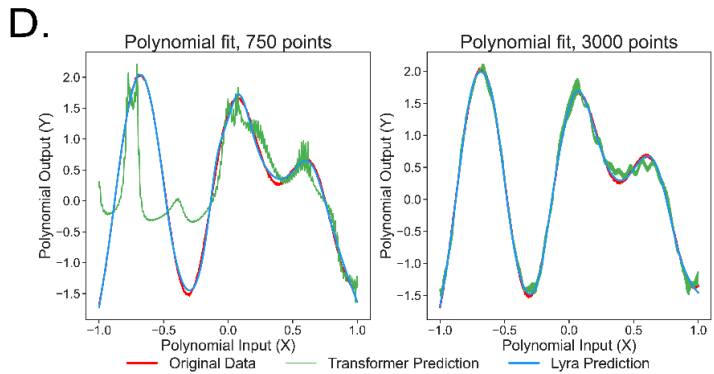

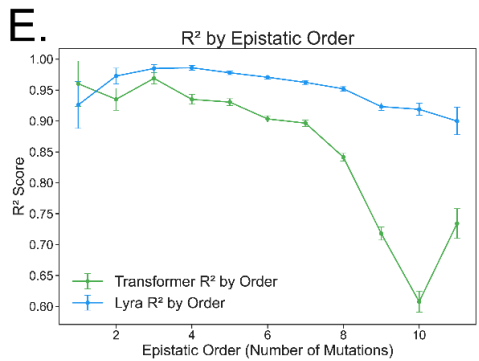
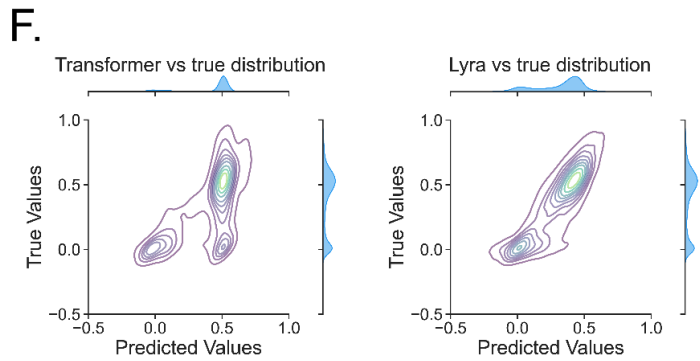

**Figure 1:** Lyra architecture enables efficient modeling of epistatic interactions through learned local and global relationships (A) Architectural overview showing protein sequence processing through projected gated convolutions (PGC) and diagonalized state space models (SSMs) known as S4D layers. (B) (left) Mathematical formulation of architectural components, including projected gated convolutions and SSM representation of signals. (right) Visualization of different types of matrices (dense, Vandermonde, Toeplitz) used in various machine learning (ML) architectures. The convolution filters produced by the Diagonalized SSM (S4D) layer of Lyra are materialized through a Vandermonde matrix, enabling the system to learn a set of basis polynomials. (C) Visualization of different S4D kernels for a system with 16 filters of length 96. (D) Comparison of polynomial approximation capabilities on synthetic data between Lyra and a similarly-sized Transformer model. (E) Regression performance across different orders of epistatic interactions (1st-11th order), demonstrating Lyra's superior ability to accurately model higher-order interactions in an entacmaea quadricolor fluorescent protein dataset[32,41]. (F) Fitness landscape visualization showing how Lyra better characterizes the distribution of protein fitness from a GFP dataset[42] compared to a similarly-sized Transformer model.

## Results

*Polynomial expressivity enables efficient mapping of biological sequences to functions*

Lyra is a novel sequence modeling architecture designed to align with the mathematical structure of intra-molecular epistatic interactions in biological sequences. Its design integrates Projected Gated Convolutions (PGCs)[43] for efficient local modeling and diagonalized SSMs (S4D) [38] to implement a circular (non-casual) convolution for capturing long-range dependencies (Figure 1A-B). By combining these components, Lyra bridges local and global sequence features, all while maintaining computational efficiency, enabling a principled approach to modeling the inherent syntactic and functional structure of biological sequences.

Epistasis arises whenever the contribution of a sequence element $u_i$ depends on other positions $u_j$ [32–35]. An epistatic landscape of a sequence $u_1 \ldots u_N$ with epistatic order $K$ can be written as

$$f(\mathbf{u}) = \sum_{k=1}^{K} \sum_{1 \leq i_1 < i_2 < \cdots < i_k \leq N} c_{i_1 i_2 \cdots i_k} \, u_{i_1} u_{i_2} \cdots u_{i_k}$$

where the learned coefficients $c_{i1,\ldots,iN}$ capture higher-order interactions[44]. Directly solving for all possible epistatic interactions is infeasible for higher orders $k$ and large sequence lengths N, motivating an architecture that *implicitly* approximates these polynomial interactions.

Lyra addresses local epistatic effects through a PGC. First we transform through a projection that makes the features richer. The transformed sequence is then processed through two parallel pathways: one applies a depthwise 1D convolution to extract local dependencies, while the other uses a linear projection to modulate how the convolution learns relationships. This gating of two layers explicitly encodes second-order interactions. By stacking such layers, Lyra can capture even higher-order dependencies without explicitly enumerating them. Further derivations, including a detailed expansion of these multiplicative terms, can be found in the supplemental.

Building on the PGC's local processing, Lyra captures longer-range dependencies using SSMs, which we demonstrate theoretically and validate empirically to be efficient polynomial approximators. Originally developed to model dynamical systems, an SSM in discrete time evolves a hidden state $x_t$ under a linear difference equation (Figure 1B):

$$x_{t+1} = \mathbf{A}\, x_t + \mathbf{B}\, u_t, \quad y_t = \mathbf{C}\, x_t,$$

where $u_t$ is the input and $y_t$ the output.

Lyra adopts a *diagonalized* state-space model, specifically S4D[38], offering computational efficiency and expressive modeling power. Central to our insight is recognizing that linear SSMs implicitly structure their hidden states to approximate polynomials characterizing sequence dynamics, a perspective closely aligned with our formulation of epistatic interactions as multilinear polynomials. S4D's expressivity emerges explicitly from the initialization of its diagonal *A* matrix, whose imaginary parts control filter frequencies while the real parts dictate filter decay (Figure 1B). Additionally, the structured initialization of the SSM's matrices naturally supports efficient convolution kernel computation, leveraging the inherent Vandermonde structure that emerges directly from the diagonalized A matrix (Figure 1B). A detailed theoretical exposition of insights into modeling epistatic interactions via SSMs and the implementation of Lyra can be found in the supplemental.

Through these capabilities, Lyra captures the multi-scale dependencies inherent in biological epistasis, outperforming Transformer-based approaches, particularly for higher-order interactions. This alignment of architecture with biological principles enables Lyra to uncover complex relationships in fitness landscapes, providing a principled and efficient framework for sequence-based modeling tasks. For example, when tasked to learn a synthetic polynomial function, Lyra demonstrates superior approximation capabilities compared to a 30% larger Transformer model trained on identical input data (Figure 1D). This polynomial approximation capability is key to effectively modeling the higher-order dependencies that characterize epistatic interactions.

Lyra's theoretical advantages translate directly to modeling epistasis in biological applications. When tested on an Entacmaea quadricolor fluorescent protein dataset with known epistatic effects[32,41], Lyra maintains high performance even for higher-order epistatic interactions where Transformer performance degrades significantly (Figure 1E). This ability allows Lyra to accurately capture complex interaction patterns across different epistatic orders. In line with this, Lyra more faithfully recovers the bimodal distribution of protein fitness in another fluorescent protein dataset[42], whereas Transformer-based models fail to preserve this epistatic structure (Figure 1F). This improved capture of the fitness landscape demonstrates how Lyra's architectural innovations enable more faithful modeling of complex biological sequence relationships.

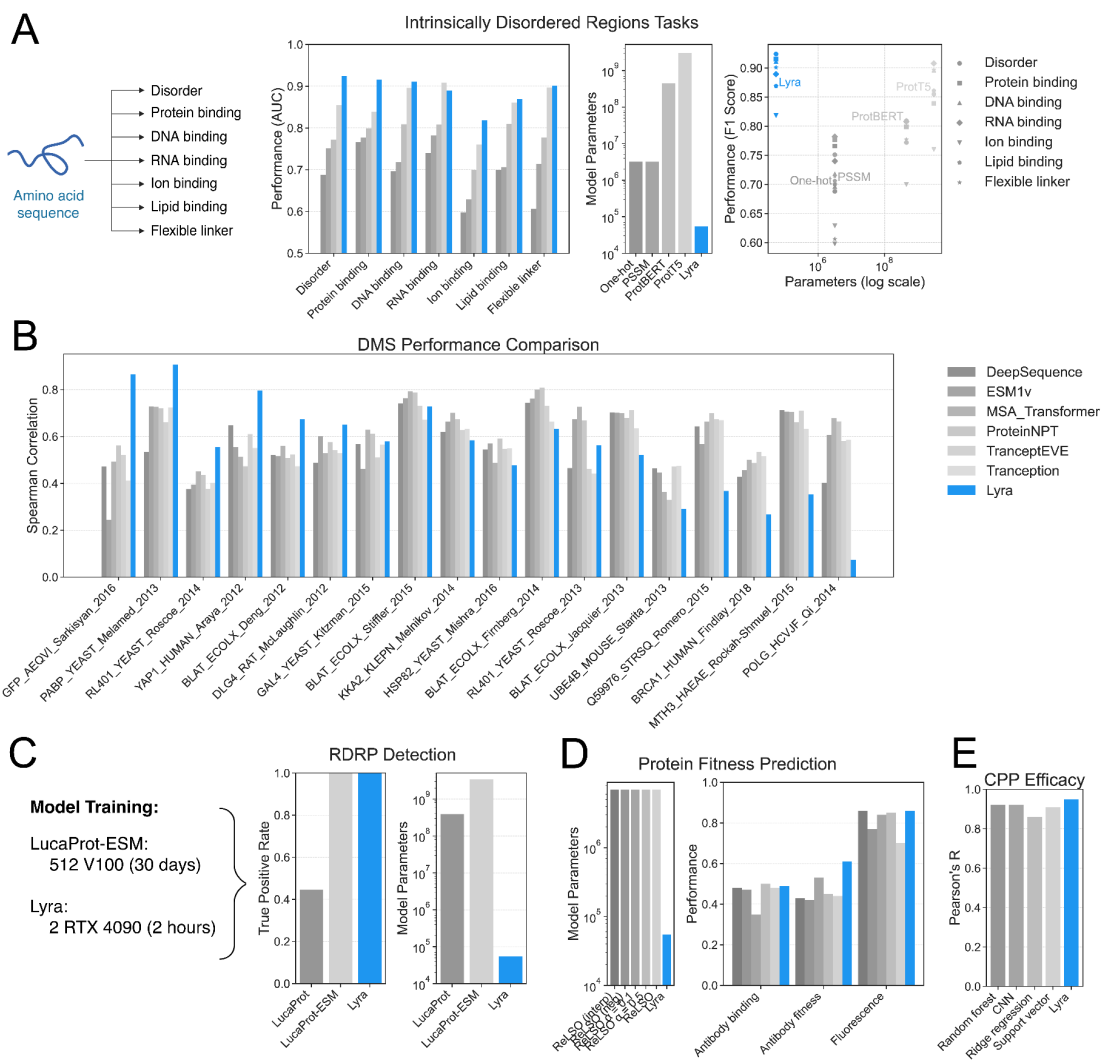

**Figure 2:** Lyra achieves state-of-the-art (SOTA) performance across diverse protein prediction tasks. **(A)** Schematic of intrinsically disordered protein regions prediction tasks. Performance comparison on disorder prediction tasks as conducted by Pang *et al* [45], demonstrating Lyra performance compared to models using input position-specific scoring matrices, one-hot encoding, or protein-language based feature representations ProtT5 [46] and ProtBERT [47]. **(B)** Performance of Lyra on deep mutational scanning (DMS) tasks[48,49] compared to baseline models[24,50–53] across multiple protein families. Lyra achieves SOTA accuracy with a significantly smaller parameter count, highlighting its efficiency in mutation effect prediction. **(C)** Lyra achieves SOTA detection accuracy for RNA-dependent RNA polymerases (RDRPs) with significantly lower computational requirements. Performance is compared to baseline models using sequence-based and structure-aware versions of LucaProt[54]. **(D)** Lyra accurately predicts protein fitness landscapes for antibody tasks and GFP brightness[55]. Performance is benchmarked against existing models, demonstrating Lyra's ability to capture complex sequence-function relationships. **(E)** Lyra achieves state-of-the-art regression performance on the Pentelute cell-penetrating peptide (CPP) dataset[56], surpassing previous models in accuracy while maintaining computational efficiency.

*Efficient protein function modeling across diverse functional landscapes*

Understanding how protein sequences encode biological function remains a central challenge in molecular biology. This relationship is inherently complex due to epistasis - where the effect of one amino acid depends on the presence of other amino acids throughout the protein sequence [35,41]. These higher-order dependencies create intricate fitness landscapes that have traditionally required either highly specialized architectures for specific tasks or massive pretrained models capturing broad sequence patterns [1,7,8,16,57,58]. While specialized models can achieve high accuracy on individual tasks and large models can generalize across functions, both approaches face significant computational limitations, especially when analyzing longer sequences or larger datasets[28,37,59] .

We benchmarked Lyra's architectural innovations against current SOTA protein function prediction methods across diverse tasks, comparing both performance and computational requirements. To ensure robust and fair evaluation of model performance, we adhered to predefined train/test splits where available, as documented in the original datasets. For datasets without predefined splits, we applied prescribed partitioning methods or, where unavailable, used random splits that maintained the original data distribution. This consistent approach ensures comparability across tasks and avoids potential biases in data handling. Further details for each task, including model architecture and training regimes, are provided in the Methods section.

Lyra enables SOTA prediction of intrinsically disordered protein regions, which represent a crucial aspect of protein function[60,61]. These regions, which lack stable 3D structure, play essential roles in cellular signaling and are implicated in neurodegenerative diseases. Notably, the Alzheimer's-associated Amyloid-β and Parkinson's-associated α-synuclein proteins are intrinsically disordered[60,62,63]. In terms of performance, Lyra achieves SOTA accuracy in six out of seven tasks, including accuracies of 0.931, 0.925, and 0.935 for disorder, protein binding, and DNA binding predictions respectively, compared to ProtT5's 0.855, 0.839, and 0.896 for the same tasks[46]. While ProtT5 required pre-training on 14 billion amino acids using 5,616 GPUs, Lyra is trained solely on the task-specific dataset of 300,000 amino acids using two GPUs in 10 minutes, using only 56,000 parameters compared to ProtT5's approximately 3,000,000,000 parameters - a >50,000-fold reduction in parameter count (Figure 2A).

Lyra achieves SOTA performance on more deep mutational scanning (DMS) tasks in our evaluation than any other tested model, ranking first in 6 out of 18 datasets (Figure 2B). DMS systematically measures how mutations affect protein function, making it a critical benchmark for assessing sequence-function models. While ProteinNPT[51] had the highest overall average $R^2$ (0.608 vs. 0.549 for Lyra), Lyra exhibited notably larger performance margins in the tasks where it led. Specifically, across its six SOTA datasets, Lyra surpassed the next-best models by an average margin of 0.150—substantially higher than margins observed for MSA Transformer (0.023 across five tasks)[24], ProteinNPT[51] (0.017 across three tasks), and TranceptEVE[52] (0.014 across two tasks). Notably, ESM1v[50] did not achieve SOTA on any DMS task. This suggests that Lyra is uniquely suited to certain mutational landscapes, excelling in diverse scenarios where other models exhibit limited improvement. Lyra's SOTA performance included tasks such as enzyme activity (BLAT_ECOLX), RNA-binding proteins (PABP_YEAST), and fluorescent proteins (GFP_AEQVI). These predictions are achieved with a 55,000-parameter Lyra model, whereas all comparison models are larger than 80M parameters, more than a 1,500x reduction in parameter count.

Lyra enables SOTA detection of RNA-dependent RNA polymerases (RDRPs), highly conserved viral proteins crucial for identifying novel pathogens and advancing our understanding of viral biology and evolution[64–67]. Using sequence information alone, Lyra achieves a 0.998 true positive rate in detection tasks, equalling LucaProt-ESM's performance when LucaProt-ESM incorporates structure-aware ESMfold embeddings, and more than doubling LucaProt's structure-free variant (0.445 true positive rate)[54]. This performance was achieved by training Lyra from scratch within 2 hours on two RTX 4090 GPUs, compared to LucaProt-ESM's compute requirements: 512 V100 GPUs for 30 days to train the ESMfold foundation model[16], followed by task-specific fine-tuning on A100 GPUs (Figure 2C). Notably, Lyra uses the same 56,000 parameter model architecture here as in the intrinsically disordered protein task, compared to 3.39 billion parameters in LucaProt-ESM, a >60,000-fold reduction in parameter count.

Lyra enables SOTA prediction of mutation impacts on protein function, demonstrated through antibody performance and fluorescent protein brightness prediction tasks (Figure 2D). Predicting how sequence changes affect protein function is essential for protein engineering[11,68–70], with antibodies serving as sophisticated test cases due to their complex sequence-function relationships[71–73] and GFP providing a well-characterized system for validating epistatic effects[42,44,74]. While ReLSO, a Transformer-based model designed for protein fitness landscape prediction [55], applies latent space optimization to refine sequence-function relationships, Lyra achieves comparable or superior performance across multiple tasks, performing competitively in antibody binding (0.49 vs. 0.50), matching it on GFP brightness (0.86 vs. 0.86), and surpassing

RELSO variants in stability prediction (0.61 vs. 0.53). Notably, while there are five RELSO variants, none achieves top-2 performance in more than one task, whereas Lyra ranks first or second across all three tasks. These predictions are achieved using a 55,000 parameter Lyra model, compared to RELSO's 7,000,000 parameters (Figure 2D) – a >120x reduction in parameters.

Lyra delivers SOTA performance in predicting cell-penetrating peptides (CPPs), which are essential for transporting therapeutic cargo across cellular membranes and play a crucial role in drug delivery (Figure 2E). Using CPP data from *Pentelute et al.[56]*, Lyra achieves a Pearson's correlation of 0.95 in a regression task, outperforming the previous SOTA of 0.92 set by a nonparametric random forest model. The model maintains the same efficient 56,000-parameter architecture used in other tasks, enabling rapid prediction of CPP effectiveness.

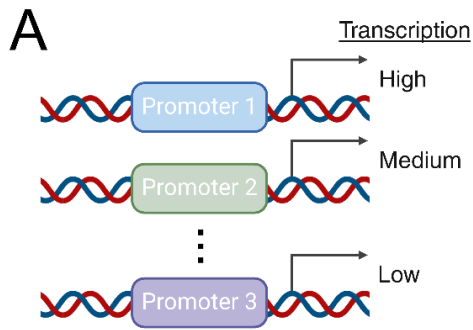
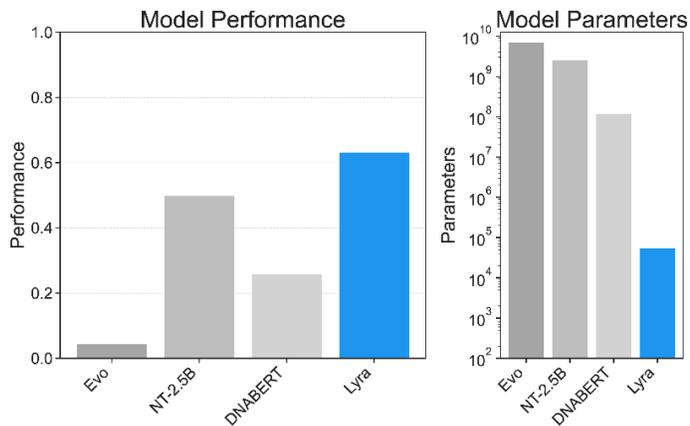
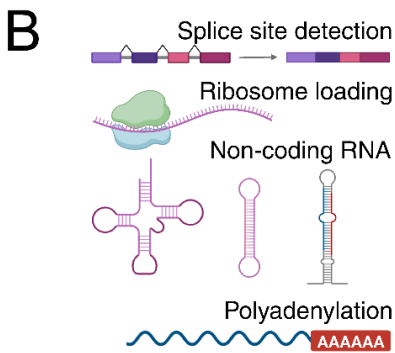
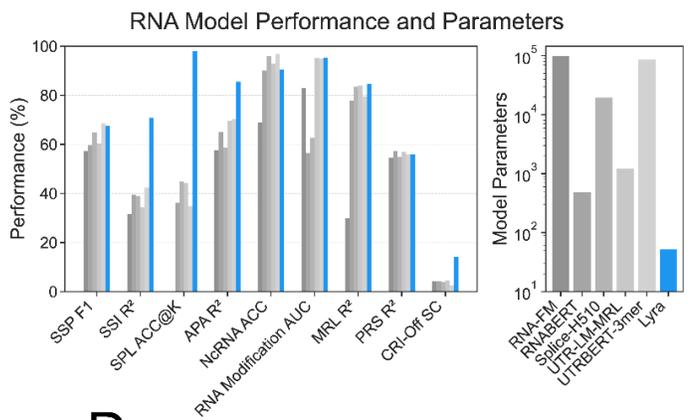
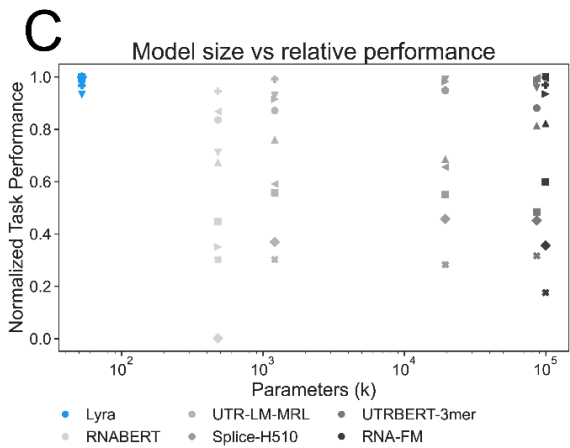
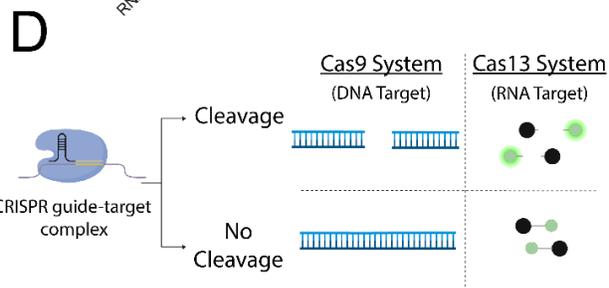
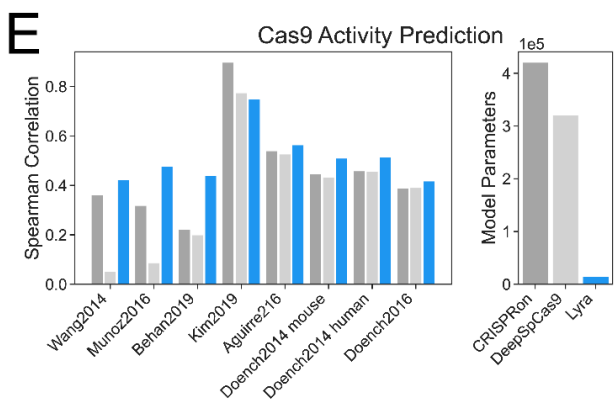
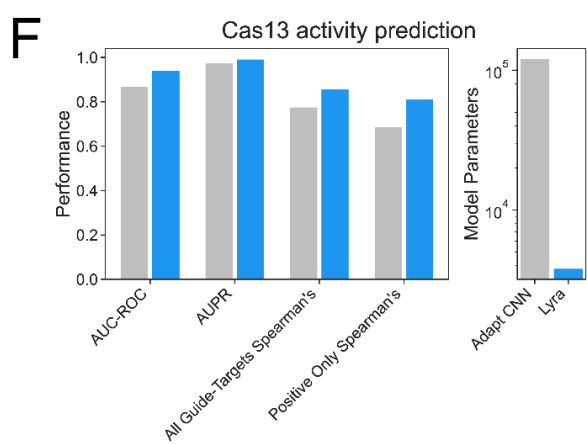

**Figure 3:** Lyra performance on DNA and RNA sequence analysis tasks. **(A)** (left) Schematic of promoter strength prediction, showing how sequence variations influence transcription levels. (right) Model parameter count and performance comparison for promoter activity prediction. **(B)** Overview of RNA prediction tasks: splice site detection, ribosome loading efficiency, non-coding RNA classification, and polyadenylation site prediction. Comparative analysis across RNA-FM[75], UTRBERT-3mer[14], SpliceBERT-H510[76], UTR-LM-MRL[77], RNABERT[78], and Lyra models. These models were selected as the best-performing representatives from different model families tested in BEACON[79]. **(C)** Comparison of relative model performance (where the best performing model in a task is normalized to 1) with respect to model size. Here, Lyra (blue) is compared to the best performing models (greys) of different parameter size ranges. Secondary structure prediction – ○; Structural score imputation – □; Splice site prediction – ◊; APA Isoform Prediction – △; noncoding RNA classification – ▽; RNA Modification – ◁; Mean ribosome loading – ▷, Programmable RNA switches ⬟, CRISPR-Off target rate prediction – ✕. **(D)** Schematic of CRISPR Cas9 and Cas13 cleavage. **(E-F)** Comparison of Lyra to other CRISPR guide-target prediction tools, including CRISPRon and DeepSpCas9 for Cas9 prediction and Adapt for Cas13 prediction.

*Efficient nucleic acid modeling spans regulatory, structural, and engineering tasks*

Understanding RNA and DNA sequence function is another fundamental challenge in molecular biology, requiring models that can capture both local motifs and long-range dependencies[2,12,80–82]. These sequences control gene expression through diverse mechanisms including promoter activity, splice site selection, and translation regulation[79,81–83]. Additionally, the growing field of gene editing relies on precise protein-RNA-DNA interactions, where RNA guides must accurately target specific DNA sequences[19,84]. Traditional architectures have struggled to simultaneously capture these varied functional elements while maintaining computational efficiency[2,85,86].

Lyra sets SOTA performance in predicting promoter activity levels. Promoter sequences serve as critical control switches for gene expression, determining where and how strongly genes are activated in cells, with accurate prediction being essential for understanding gene regulation and designing synthetic genetic circuits[13,81,87]. In performance testing across prokaryotic promoters, Lyra achieves a Spearman correlation of 0.63, substantially outperforming modern foundation models (Nucleotide Transformer-2.5B[23]: 0.50, DNABERT[88]: 0.26, Evo[8]: 0.04). Remarkably, a Lyra model accomplishes this with only 54,000 parameters, contrasting sharply with larger models like Evo (7 billion parameters), NT-2.5B (2.5 billion parameters), and DNABERT (117 million parameters), representing more than a 120,000-fold reduction in parameters while improving performance (Figure 3A).

Lyra achieves SOTA performance on the BEACON RNA benchmarking suite[79] (Figure 3B). This comprehensive benchmark evaluates RNA sequence analysis across distinct tasks critical for understanding gene regulation, from splice site recognition to ribosome loading prediction. In performance testing, Lyra sets new SOTA in five out of the nine tasks that we tested, with particularly dramatic improvements in structural score imputation ($r^2$ of 0.7305 versus previous best RNA-FM's[75] 0.4236) and splice site prediction (98.89% accuracy versus previous best Splice-MS510's[89] 50.55%, a relative improvement of 95.63%). Lyra is performant across all tasks; in the four tasks where it does not set SOTA, its relative performance is within 7% of the best result (Figure 3C). In contrast, the next best model, UTRBERT-3mer[14], is on average 23.17% below SOTA across tasks and is 66.81% below Lyra's performance in its worst task. These improvements are achieved using Lyra models ranging from 46,000 to 58,000 parameters, compared to 86.1 million parameters for UTRBERT-3mer and up to 99.5 million parameters for RNA-FM (Figure 3B), up to a 2100-fold reduction in parameters.

Finally, we examine Lyra's performance and computational efficiency for studying CRISPR guide RNA activity prediction (Figure 3D). Accurate guide prediction is crucial for both diagnostic and therapeutic applications, where guide-target recognition efficiency can vary by orders of magnitude (Figure 3F)[84,90,91]. These applications span from Cas9-mediated genome editing needing efficient DNA targeting to Cas13-based viral diagnostics requiring precise RNA detection (Figure 3C).

Lyra sets SOTA performance in Cas9 genome editing prediction accuracy. Precise genome editing requires guides that efficiently and specifically direct Cas9 to target DNA sequences, with guide selection directly impacting therapeutic outcomes[90,91]. Evaluated across eight independent datasets spanning different experimental conditions and cell types[90,92–97], Lyra achieves an average Spearman correlation of 0.51 compared to 0.45 and 0.36 for CRISPRon[19] and DeepSpCas9[95], improving performance in seven out of eight datasets. Notably, Lyra more than doubles prediction accuracy on the challenging Behan2019 dataset[93] with a correlation of 0.439 versus 0.219 and 0.198 for CRISPRon and DeepSpCas9. These advances are achieved using just 14,000 parameters compared to CRISPRon's 420,000 and DeepSpCas9's 320,000 parameters (Figure 3D), more than a 20-fold reduction in parameters.

Lyra demonstrates SOTA performance in Cas13-based diagnostic applications. These systems require highly specific guide-target recognition for reliable viral detection and pathogen surveillance[84,98,99].

In classifying between active and non-active guide-target pairs, Lyra achieves an AUC-ROC of 0.939 and AUPR of 0.990, outperforming the previous SOTA ADAPT model (0.866 and 0.972). For quantitative activity prediction, Lyra achieves Spearman correlations of 0.856 and 0.810 for all guide-target pairs and positive-identified pairs respectively, compared to ADAPT's 0.774 and 0.686. These improvements are achieved with just 3,800 parameters, a >30-fold reduction from ADAPT's 120,000 parameters (Figure 3E).

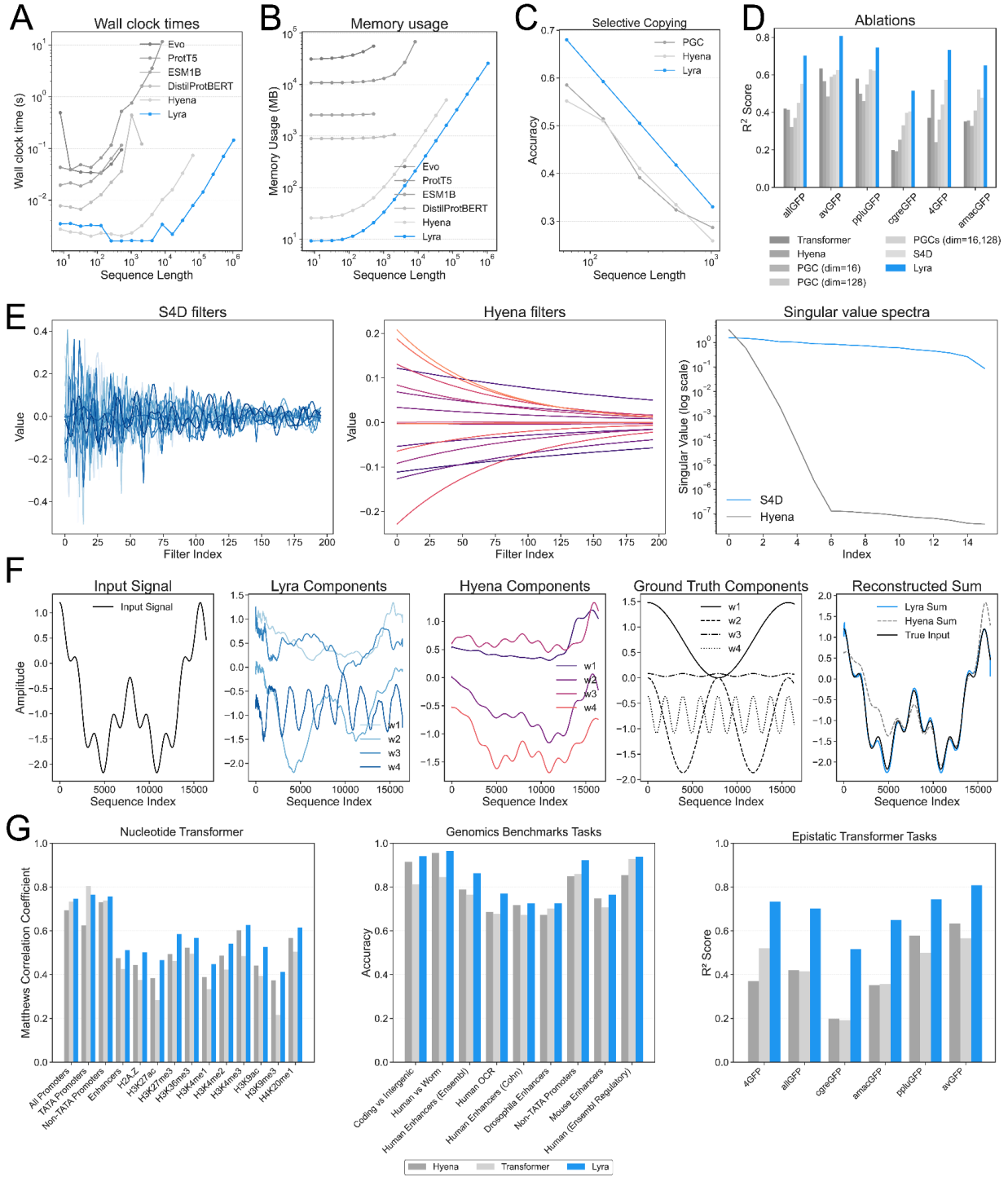

**Figure 4:** Computational efficiency analysis of Lyra compared to existing models. **(A)** Wall clock time versus sequence length, demonstrating Lyra's favorable scaling compared to Transformer-based (ESM-1b[9], DistilProtBert) and other convolutional (Hyena) architectures. **(B)** Memory requirements comparison across models, highlighting Lyra's reduced resource needs. **(C)** Performance on the selective copying synthetic task, where models are evaluated on their ability to identify

mutations occurring at non-uniform intervals. The task is based on GFP sequence mutations, and models are assessed using accuracy metrics. **(D)** Ablations done by removing different components of Lyra on GFP tasks, showing comparisons between PGC-only models, S4D-only models, and Lyra, along with comparisons with Transformer and Hyena-based models. **(E)** Visualization of convolution filters in S4D and Hyena, including an investigation of singular values across different sequence modeling primitives. **(F)** Frequency analysis task comparing Lyra and Hyena on reconstructing a composite sequence consisting of four weighted cosine waves. Lyra accurately recovers frequency components, demonstrating superior capability to separate overlapping signals relevant to interpreting biological sequences. **(G)** Benchmarking results across regulatory genomics (left, middle) and proteomics tasks (right), with comparisons to equivalently-sized Hyena and Transformer models.

*Improved inference speed and memory requirements lowers barriers in biological modeling*

Beyond setting new SOTA benchmarks across diverse tasks, Lyra reduces computational requirements, enabling more widespread adoption. With O(N log N) computational scaling relative to sequence length—compared to the quadratic complexity of attention mechanisms[21]—Lyra achieves substantial speedups across various sequence lengths and batch sizes (Figure 4A). For sequence length 512, Lyra is 28.4x, 69.71x, and 239.2x faster than ESM-1b[9] for batch sizes 1, 2, and 8, respectively, and 214.2x and 56.6x faster than Evo[8] for batch sizes 1 and 2. Batch size 8 did not run for Evo due to memory constraints, and ESM-1b ran out of memory for sequence length 1,024 at batch size 1 on an Nvidia A100.

Lyra's sub-quadratic scaling also enables the processing of substantially longer sequences than Transformer-based foundation models. Due to the quadratic scaling of computation and memory, Transformer-based models become computationally infeasible in our experimental setup beyond sequence lengths of 4,096. In contrast, Lyra efficiently processes sequences up to 65,536 in length, enabling significantly longer-range modeling due to its subquadratic complexity. On an Nvidia A100, Lyra achieves this in just 7.9ms at batch size 2, making it 8.3x faster than Hyena (Figure 4A).

In addition to its speed, Lyra dramatically reduces memory requirements. Most tasks in this paper were computed with a Lyra models of approximately 55,000 parameters, which performed favorably compared to models such as ESM-1B and Evo, with 650 million[9] and 7 billion parameters[100], respectively—reductions of >10,000-fold and >120,000-fold in parameter count. These significant reductions translate directly to lower memory usage, enabling training and deployment on consumer-grade hardware. For example, at sequence length 512 and batch sizes 1, 2, and 8, Lyra uses 125.8x, 127.9x, and 138.1x less memory than ESM-1b and 2278.1x and 2661.8x less memory than Evo for batch sizes 1 and 2, respectively (Figure 4B). Due to these low memory requirements, we were able to train and run every task in this study on two or fewer GPUs in under two hours, a stark contrast to prior approaches often requiring clusters of specialized GPUs running for days or weeks.

To evaluate Lyra's ability to accurately identify biologically meaningful signals within extensive sequences, we adapted the selective copying task[31], which tests a model's capability to selectively recognize and reproduce specific tokens placed within longer sequences. Biological sequences often feature sparse yet crucial signals, such as mutations, distributed among non-informative regions. We defined a consensus (wild-type) GFP sequence, identified frequently mutated positions, and patterns of co-mutation. These biologically-informed mutations were then added at irregular intervals within synthetic sequences, while other positions contained zero tokens. Lyra consistently outperformed Hyena and PGC-only models in correctly identifying both the locations and identities of these mutations (Figure 4C), demonstrating its effectiveness in capturing biologically realistic, context-dependent signals across sequence lengths. Secondarily, we observed that Hyena and PGC performance were comparable on this synthetic task, indicating that PGCs exhibit similar expressivity to Hyena in the context of selective copying.

We performed extensive ablation studies to evaluate the contribution of each architectural component to Lyra performance (Figure 4D). Using six different GFP datasets (allGFP, avGFP, ppluGFP, cgreGFP, 4GFP, and amacGFP [44]), we compared the full Lyra architecture against individual components and baseline models. Notably, across all six datasets, the full Lyra architecture ($R^2$ from 0.52 to 0.81) consistently outperforms not only individual components like single PGC layers ($R^2$ from 0.24 to 0.59) and S4D alone ($R^2$ from 0.41 to 0.63), but also alternative architectures like Hyena ($R^2$ from 0.19 to 0.57) and Transformers ($R^2$ from 0.20 to 0.63). The improvements are particularly striking on challenging datasets like cgreGFP, where Lyra achieves an $R^2$ of 0.52 compared to 0.20 for the Transformer baseline and 0.19 for Hyena, representing a more than 2.5x improvement in performance. These results validate our architectural choices and demonstrate that the combination of PGC and S4D components enables more effective modeling of epistatic interactions than any component alone or alternative architecture.

We analyzed the convolution kernels generated by Lyra and Hyena layers to investigate their architectural difference (Figure 4E). Kernels from Lyra's S4D layers exhibited distinct oscillatory patterns characterized by decay, a direct consequence of the parameterization of the state-transition *A* matrix. Specifically, the real component of the *A* matrix determines the decay rate, while the complex component governs the oscillation frequency. This structured parameterization significantly enhanced sequence reconstruction accuracy in our synthetic experiments. In contrast, Hyena filters displayed smooth, gradually decaying curves without pronounced oscillatory features, reflecting their kernel parameterization via a multi-layer perceptron (MLP) with positional embedding. Singular value decomposition (SVD) of both kernels further indicated that Lyra's filters maintained higher rank and slower singular value decay across dimensions, implying more meaningful and broadly distributed representations within the kernels. Conversely, Hyena filters exhibited sharper singular value decay, suggesting that relevant characterizations were confined to fewer dimensions.

We evaluated Lyra's capability to reconstruct complex composite signals composed of multiple overlapping frequency components (Figure 4F), relevant to protein sequence analysis, where different frequencies correspond to different functional features. Synthetic sequences, generated as weighted sums of four distinct cosine waves with unique frequencies, coefficients, and offsets, tested models' reconstruction abilities. Lyra achieved a higher reconstruction accuracy and more accurately predicted individual frequency components than Hyena. This aligns with our investigation of Hyena and S4D filters from Figure (4E), where we observed that S4D's kernels parameterize oscillatory filters exhibiting clear decay patterns, whereas Hyena's kernels display smoother, gradually decaying curves without significant oscillatory behavior, suggesting a lower capacity for representing complex frequency dynamics compared to S4D.

To further assess Lyra's effectiveness as a general-purpose architecture, we compared it to similarly-sized Hyena-based and Transformer-based models across genomics and protein function tasks (Figure 4G). Hyena employs depthwise convolutions and gated long convolutions, while we use an optimized Transformer recipe that integrates rotary position embeddings (RoPE), RMSNorm pre-normalization, and SwiGLU for feature extraction. Additionally, we directly compared Lyra to Transformers on epistatic modeling tasks to evaluate its ability to capture complex mutation interactions.

Lyra consistently improved performance across tasks, outperforming Hyena and Transformers on 9 out of 9 Genomics Benchmark[101] tasks and 17 out of 18 Nucleotide Transformer[23] tasks. In Epistatic Transformer tasks[44], Lyra outperformed Transformers across all 27 tasks and surpassed Hyena in the 6 tasks where it was evaluated (Figure 4G, Supplemental Figure 1). For nucleotide sequence classification, Lyra excelled in regulatory sequence tasks, including promoter, enhancer, and histone modification prediction. Similarly, in genomic classification, Lyra achieved superior accuracy in functional annotation tasks such as coding vs. intergenic regions, enhancer detection, and regulatory element classification. Furthermore, in modeling epistatic interactions, Lyra outperformed other methods in capturing higher-order dependencies within biological sequences, demonstrating improved accuracy in tasks requiring an understanding of non-additive mutation effects and fitness landscapes (Supplementary Figure 1); this was especially evident in fluorescent protein datasets, where Lyra achieved the largest performance gains (Figure 4G). These results establish Lyra as a highly generalizable architecture that consistently outperforms both convolutional and Transformer-based approaches, offering a robust foundation for sequence-to-function prediction with minimal task-specific tuning.

**Discussion**

The ability of SSMs to efficiently approximate multilinear polynomial functions provides a powerful new mathematical framework for modeling biological sequences. This fundamental insight enables Lyra to capture complex epistatic interactions - where the effect of one mutation depends on other mutations - more effectively than previous approaches. By combining this mathematical foundation with PGCs for extracting local sequence features, Lyra achieves SOTA performance across diverse biological challenges while using significantly fewer parameters than established models. Lyra achieves comparable or superior performance on most tasks, training from scratch with just one or two GPUs and completing both training and task execution in minutes to hours, compared to current foundation models that require massive computational infrastructure and weeks of training time [8,16,28].

The success of a simple, mathematically-principled architecture in outperforming both large foundation models and structure-aware approaches challenges current trends in computational biology[9,27,28,102]. While large language models trained on billions of sequences have demonstrated remarkable capabilities, Lyra shows that architectural innovations informed by the underlying mathematics of biological phenomena can achieve superior results with orders of magnitude less computation. This suggests that understanding and encoding domain-specific mathematical structures may be more valuable than increasing model scale.

Lyra's performance and efficiency across diverse tasks, from protein fitness landscapes to RNA splicing, demonstrates the broad applicability of polynomial approximation in biological sequence analysis. The architecture excels particularly in capturing non-linear interactions in protein fitness landscapes, predicting complex CRISPR guide-target interactions, and modeling RNA structural features. These capabilities point to immediate applications in therapeutic development and pathogen surveillance. Specifically, the model's efficiency in capturing sequence-to-function relationships could accelerate the design of cell-penetrating peptides for drug delivery, optimize viral vehicles for targeted gene therapy, predict solubility of biologic drug candidates, and enable rapid detection of viral threats through efficient viral protein identification. Beyond therapeutics and surveillance, Lyra could enhance biomanufacturing through improved discovery and optimization of catalytic enzymes. The model's computational efficiency makes rapid iteration on these applications feasible even with limited computational resources, potentially accelerating the development pipeline from sequence design to experimental validation.

Lyra challenges the prevailing trend towards increasingly larger models in biological sequence analysis, achieving SOTA performance while democratizing access to researchers without requiring specialized computing infrastructure. By understanding and encoding the mathematical principles underlying biological phenomena - in this case, the polynomial nature of epistatic interactions - we achieve dramatically more efficient solutions. Looking forward, the connection between SSMs and polynomial approximations, as demonstrated in Lyra, may have far-reaching implications beyond biological sequences, offering a promising approach for domains where complex interactions follow polynomial-like behavior.

## Methods and Materials

*Lyra Architecture*

Lyra comprises two core components: the Projected Gated Convolution (PGC) block[43], followed by a state-space layer with depthwise convolution (S4D). In the standard implementation, which consists of approximately 55,000 parameters, Lyra includes two PGC blocks. The first PGC block operates with a hidden dimension of 16, while the second uses a hidden dimension of 128. These are followed by an S4D layer[38], which has a hidden dimension of 64 and is equipped with a residual connection and sequence prenormalization using Root Mean Square Layer Normalization (RMSNorm). The PGC blocks are designed to capture contextualized local dependencies in the input sequence, while the S4D layer parameterizes a long convolution to model long-range dependencies.

*Projected Gated Convolution (PGC)*

The PGC is the first stage of our model, designed to process biological sequences and extract both local and global features. Each layer begins by linearly projecting the input sequence, reducing or expanding its feature dimensionality to an intermediate size. This projection is followed by Root Mean Square Layer Normalization (RMSNorm[103]), which stabilizes training dynamics. The transformed sequence is then processed through two parallel pathways: one applies a depthwise 1D convolution to extract local dependencies, while the other uses a linear projection to modulate how the convolution learns relationships. These two outputs are combined using element-wise multiplication, effectively conditioning the convolution on global sequence features while preserving its local receptive field, allowing the model to learn feature interactions that are both contextually informed and positionally invariant. Finally, the combined features are projected back to the original feature dimensionality and normalized again with RMSNorm. This process allows the module to capture complex patterns and dependencies in the input sequence, making it well-suited for modeling biological data.

*S4D*

The S4D model leverages diagonal state space models (SSMs) to parameterize and compute convolution kernels for sequence modeling. The kernel, which captures dependencies across the sequence, is parameterized through three core matrices: *A*, *B*, and *C*. The matrix *A* governs the dynamics of the system, encoding exponential decay and oscillatory behavior, while *B* maps the input into the state space, and C projects the state back into the output space. The convolution kernel is computed efficiently using the Vandermonde matrix[104], which organizes the contributions of *A, B,* and *C* into a structure that allows for efficient evaluation of the kernel as a sum of weighted exponential terms.

*Biological datasets*

*Intrinsically Disordered Protein Tasks*
For predicting intrinsically disordered regions (IDRs), we utilized the dataset from [45], which contains 925 protein sequences with experimentally validated disorder annotations. The dataset was split by Peng et al into training (589 sequences), validation (148 sequences), and test (188 sequences) sets using CD-HIT clustering[105] with 25% sequence similarity threshold. The input sequences were one-hot encoded and the model was trained to predict six different types of disorder functions: protein-binding, DNA-binding, RNA-binding, ion-binding, lipid-binding, and flexible linker regions. The Lyra internal model dimension was 64, with two consecutive PGC layers with dimensions 16 and 128, respectively. Training was conducted for 30 epochs using AdamW optimizer with a learning rate of 0.001 and weight decay of 0.01.

*Deep Mutational Scanning (DMS) tasks*
Lyra was evaluated on DMS tasks using datasets from ProteinGym[53], originally curated in [49]. The datasets cover diverse mutational landscapes, including enzyme activity, RNA-binding, and fluorescent protein function. We retained ProteinGym's train-test splits. The Lyra internal model dimension was 64, with two consecutive PGC layers with dimensions 16 and 128, respectively. Lyra was trained for 100 epochs using AdamW (learning rate 0.001, weight decay 0.01) and evaluated using Spearman's rank correlation to compare predicted and experimental fitness scores. Results were benchmarked against ProteinGym baselines.

*RDRP Tasks*
The RNA-dependent RNA polymerase (RDRP) detection task utilized the dataset from [54], focusing on identifying RDRP sequences from amino acid inputs. This binary classification task involved classifying whether a particular input amino acid sequence represents an RDRP. The dataset was used as split by [54] intro training/validation/test splits, with 5,979 total positive viral RDRP sequences and 150,000 sequences sampling from negative samples. The Lyra internal model dimension was 64, with two consecutive PGC layers with dimensions 16 and 128, respectively. Training was performed for 100 epochs using the AdamW optimizer with a learning rate of 0.001 and weight decay of 0.01. Model performance was evaluated primarily using the true positive rate metric to enable direct comparison with the LucaProt baseline, alongside standard binary classification metrics including accuracy and AUC-ROC.

*ReLSO Tasks*

For the protein fitness prediction tasks, Lyra was trained across three fitness prediction datasets GB1[106], Gifford[107], and GFP[108]. Each dataset contained amino acid sequences of the same length which were one-hot-encoded, input dimension of 20, with the stability and affinity, enrichment, or fluorescence values serving as regression labels . The train/test split was as given from the source datafiles in the ReLSO github repository. The Lyra internal model dimension was 64, with two consecutive PGC layers with dimensions 16 and 128, respectively. The training was performed for 100 epochs, utilizing the AdamW optimizer with a learning rate of 0.001 and a weight decay of 0.01. The evaluation metric was Spearman's rank correlation coefficient on the validation set, and Mean Squared Error Loss (MSELoss) was used as the loss function.

*Cell-Penetrating Peptides Efficacy*

We evaluated performance on cell-penetrating peptide (CPP) efficacy prediction using the dataset from [56], which contains 640 sequences with experimentally validated cell penetration abilities. The models were trained for 100 epochs with a random 80/20 train/test split as conducted by the source paper. The Lyra internal model dimension was 64, with two consecutive PGC layers with dimensions 16 and 128, respectively, utilizing AdamW optimization with a learning rate of 0.001 and weight decay of 0.01. CPP efficacy was assessed using regression metrics including Spearman's correlation coefficient and MSE loss.

*CRISPR Cas13a (ADAPT)*

For the CRISPR Cas13 dataset [84], we encoded guide-target pairs using a one-hot encoding scheme with a dimensionality of 4 for each guide and target. These were then concatenated to form a stacked representation with an 8-dimensional one-hot-encoded vector for sequences of 48 base pairs. The log fluorescence threshold to distinguish active from non-active pairs was set at a value of -4.00 in the original ADAPT paper. The Lyra internal model dimension was 16, with a PGC layer with dimension 16. Our model underwent 5-fold cross-validation across three distinct tasks. In the first task, binary classification of guide-target pairs was performed, assessing the model's performance through AUC-ROC and AUPR metrics, with each fold being trained for 75 epochs. The following two tasks involved regression analyses: the first was a positive-only regression targeting values above the activity threshold, and the second encompassed a comprehensive regression across all guide-target pairs, both positive and negative. Both regression tasks were evaluated using Spearman's coefficient, following the same 75-epoch, 5-fold cross-validation structure

*CRISPR Cas9*

We utilized eight CRISPR Cas9 datasets—Kim2019[95], Doench2014 mouse[96], Doench2014 human[96], Doench2016[90], Wang2014[92], and Munoz2016[94], Aguierre2016[97], and Behan2019[93]—comprising guide-target activity information. Each sequence was one-hot encoded to capture the nucleotide arrangement. The Lyra internal model dimension was 48 with a PGC layer with dimension 128. For the purposes of model training and validation, we adhered to a 5-fold cross-validation procedure, applied to both training and test sets. Each fold was trained for 150 epochs of training, and evaluated using Spearman's correlation for regression enzymatic activity based on a sequence.

*Promoter Tasks*

The promoter strength prediction task utilized the dataset from[109], which consists of 3,665 synthetic modifications of the Ptrc promoter, engineered and characterized through iterative mutation-construction-screening cycles. This regression task aimed to predict promoter strength based on sequence inputs, with fluorescence/OD600 measurements serving as the target variable. The dataset was randomly split into a training set (90%) and a test set (10%). A small Lyra model was used, with internal model dimension 64 and PGC layers with dimension 16, and 64, totalling 54,145 parameters. Training was conducted for 100 epochs using AdamW optimizer with a learning rate of 0.001 and weight decay of 0.01.

*RNA Tasks (BEACON)*

For standard benchmarking in RNA tasks, we utilized datasets provided in the BEACON dataset by [79]. For all tasks, we used the same metric (F1 score, $R^2$, accuracy, ACC, AUC, or Spearman's) as in the BEACON manuscript to compare Lyra performance with the models tested in BEACON.

Secondary Structure Prediction

For secondary structure prediction, we utilized the bpRNA dataset[110], which provides detailed annotations of 13,419 RNA structures. Each RNA sequence is associated with a target string $y \in R^L$, indicating nucleotide pair information as part of the secondary structure. The data was split into train, validation, and testset using the provided split from the BEACON repository. The Lyra internal model dimension was 72, with two consecutive PGC layers with dimensions 18 and 144, respectively. The training protocol involved 100 epochs using the AdamW[111] optimizer with a learning rate of 0.001 and weight decay of 0.01. The task was evaluated in the RNA Beacon and in our manuscript paper using the F1 score.

*Structural Score Imputation*

The icSHAPE HEK293 dataset[112] was used for structural score imputation, containing experimentally derived RNA structural scores. The dataset consists of 14,049 training, 1,756 validation, and 3,095 test fragments. The Lyra internal model dimension was 64, with two consecutive PGC layers with dimensions 16 and 128, respectively. The training protocol involved 100 epochs using the AdamW optimizer with a learning rate of 0.001 and weight decay of 0.01. The task was evaluated in the RNA Beacon paper using and in our manuscript using the $R^2$ metric, quantifying the accuracy of imputed scores.

*Splice Site Prediction*

For splice site classification, we employed the SpliceAI dataset[113] containing 144,628 training, 18,078 validation, and 16,505 test sequences. Each nucleotide was labeled as an acceptor (a), donor (d), or neither (n), with predictions evaluated using Top-k accuracy. The Lyra internal model dimension was 72, with two consecutive PGC layers with dimensions 18 and 144, respectively. The training protocol involved 100 epochs using the AdamW optimizer with a learning rate of 0.001 and weight decay of 0.01. The task was evaluated in the RNA Beacon paper and in our manuscript using the F1 score.

*APA Isoform Prediction*

For APA isoform prediction, we employed the APARENT dataset[114] containing 145,463 training, 33,170 validation, and 49,755 test sequences. Each sequence was labeled with the usage ratio of the proximal polyadenylation site (PAS) in the 3' untranslated region (3' UTR), recorded as $y \in R$. The Lyra internal model dimension was 64, with two consecutive PGC layers with dimensions 16 and 128, respectively. The training protocol involved 100 epochs using the AdamW optimizer with a learning rate of 0.001 and weight decay of 0.01. The task was evaluated in the RNA BEACON paper using the $R^2$ metric.

*Noncoding RNA Classification*

For noncoding RNA classification, we employed the Noorul's dataset[115] containing 5,679 training, 650 validation, and 2,400 test sequences. Each sequence was categorized into one of thirteen labels, including microRNAs (miRNAs), long noncoding RNAs (lncRNAs), and small interfering RNAs (siRNAs), with labels $y \in \{0, 1, ..., 12\}$. The Lyra internal model dimension was 72, with two consecutive PGC layers with dimensions 18 and 144, respectively. The training protocol involved 100 epochs using the AdamW optimizer with a learning rate of 0.001 and weight decay of 0.01. The task was evaluated in the RNA BEACON paper and in our manuscript using the accuracy metric.

*RNA Modification Prediction*

For RNA modification prediction, we employed the MultiRM dataset[36] containing 304,661 training, 3,599 validation, and 1,200 test sequences. Each nucleotide was labeled with one of 12 different modification types, with labels y ∈ {0, 1, ..., 11}. The Lyra internal model dimension was 72, with two consecutive PGC layers with dimensions 18 and 144, respectively. The training protocol involved 100 epochs using the AdamW optimizer with a learning rate of 0.001 and weight decay of 0.01. The task was evaluated in the RNA BEACON paper and in our manuscript using the AUC metric.

*Mean Ribosome Loading Prediction*

For mean ribosome loading prediction, we employed the Optimus dataset[116] containing 76,319 training, 7,600 validation, and 7,600 test sequences. Each sequence was labeled with an MRL value y ∈ R, representing the level of mRNA translation activity into proteins. The internal model dimension was 64, with two consecutive PGC layers with dimensions 16 and 128, respectively. The training protocol involved 100 epochs using the AdamW optimizer with a learning rate of 0.001 and weight decay of 0.01. The task was evaluated in the RNA BEACON paper and in our manuscript using the $R^2$ metric.

*Programmable RNA Switch Prediction*

For programmable RNA switch prediction, we employed the Angenent-Mari dataset[117] containing 73,227 training, 9,153 validation, and 9,154 test sequences. Each sequence was labeled with ON, OFF, and ON/OFF activity states, recorded as y ∈ $R^3$. The internal model dimension was 72, with two consecutive PGC layers with dimensions 18 and 144, respectively. The training protocol involved 100 epochs using the AdamW optimizer with a learning rate of 0.001 and weight decay of 0.01. The task was evaluated in the RNA BEACON paper and in our manuscript using the $R^2$ metric.

*CRISPR Off-Target Rate Prediction*

For CRISPR off-target prediction, we employed the DeepCRISPR dataset[118] containing 14,223 training, 2,032 validation, and 4,064 test sequences. Each sequence was labeled with an off-target frequency score y ∈ R, quantifying CRISPR-induced mutations at unintended genomic locations. The internal model dimension was 62, with two consecutive PGC layers with dimensions 16 and 128, respectively. The training protocol involved 100 epochs using the AdamW optimizer with a learning rate of 0.001 and weight decay of 0.01. The task was evaluated in the RNA BEACON paper using the weighted Spearman correlation.

*Architecture Comparison Studies*

To evaluate Lyra's performance against similarly-sized, non-task-specialized architectures—including long convolutional and Transformer models—we conducted direct comparisons with Hyena-based[40] and Transformer++-based [31] models in side-by-side studies.

*Hyena*

The Hyena architecture combines short depthwise convolutions and linear projections, which are gated together with a long convolution. Unlike S4D, which uses state space models (SSMs) to parameterize input-dependent long convolution kernels, Hyena[40] employs a multi-layer perceptron (MLP[119]) with positional encoding. The Hyena model has a hidden dimension of 64, an input dimension of 1, and an output dimension of 4, totaling 47,033 parameters. Determined via Hyperparameter sweep, weight decay for the Hyena layer is set to 0, embedding dropout to 0.2, and residual dropout is configured to 0.2.

*Transformer*

The baseline is an optimized transformer recipe called Transformer++ [31], which is a Transformer model configured with rotary position embeddings (RoPE)[120], pre-normalization using root mean square layer normalization (RMSNorm), and SwiGLU[121] for dimensional mixing. In this setup, the hidden dimension of SwiGLU is four times the model width, providing the necessary capacity for feature extraction.

*Epistastic Transformer GFP tasks*

In this study, we conducted a head-to-head comparison of similarly sized Transformer models and Lyra to evaluate their capacity for modeling epistatic interactions. We focused on six GFP tasks[42] identified by [44] as being highly influenced by epistatic interactions. The datasets included different GFP variants and varied in size, sequence length, and median Hamming distance, and included preset train/test splits. Transformer and Lyra models were trained for 50 epochs, utilizing the AdamW optimizer with a learning rate of 0.001 and a weight decay of 0.01. The primary evaluation metric was $R^2$, computed for each dataset.

*Nucleotide Transformer Tasks*

We evaluated our model across 14 genomic prediction tasks from [23], which were selected to assess its ability to generalize across key regulatory and chromatin-related functions. These tasks included promoter prediction (promoter_all, promoter_tata, promoter_no_tata), enhancer prediction (enhancers), and histone modification state classification across multiple chromatin marks (H2AFZ, H3K27ac, H3K27me3, H3K36me3, H3K4me1, H3K4me2, H3K4me3, H3K9ac, H3K9me3,

H4K20me1). The Lyra model had an internal model dimension of 64, consisting of two consecutive PGC layers with dimensions of 16 and 128, respectively, followed by an S4D block with a hidden dimension of 64, resulting in a total of 46,210 parameters. The Transformer++ model had a total of 48,610 parameters, with a model dimension of 64, input dimension of 4, and a SwiGLU intermediate dimension of 192; the output dimension was adjusted according to task requirements. The Hyena model contained 47,033 parameters, with a hidden dimension of 64, input dimension of 4, zero weight decay specifically applied to Hyena layers, and dropout rates of 0.2 for both embedding and residual connections. The Hyena, Transformer++, and Lyra models were trained for 50 epochs and evaluated across the Nucleotide Transformer downstream task benchmark. Tasks included both binary (2-class) and 3-class classification, with enhancer_types and splice_sites_all specifically requiring 3-class outputs. The datasets covered diverse sequence lengths ranging from 300bp to 1,000bp.

*Genomics Benchmarks*

We also evaluated our model using the Genomics Benchmark [101] dataset, which is a curated collection of publicly available genomic classification tasks designed for benchmarking deep learning models. For our evaluation, we selected nine tasks spanning key functional genomics challenges, including sequence classification (Coding vs Intergenic, Human vs Worm), enhancer and promoter prediction (Human Enhancers (Ensembl), Human OCR, Human Enhancers (Cohn), Drosophila Enhancers, Non-TATA Promoters, Mouse Enhancers), and regulatory element classification (Human Ensembl Regulatory). The Lyra model had an internal model dimension of 64, composed of two consecutive PGC layers with dimensions of 16 and 128, respectively, followed by an S4D block with a hidden dimension of 64, resulting in a total of 46,210 parameters. The Transformer++ model had a total of 48,610 parameters, featuring a model dimension (d_model) of 64, input dimension (d_input) of 4, a SwiGLU intermediate dimension of 192, output dimensions of either 2 or 3 depending on the specific task, weight decay to 0, embedding dropout of 0.2, residual dropout of 0.2, and 8 attention heads. The Hyena model contained 47,033 parameters, featuring a hidden dimension of 64, input dimension of 4, output dimensions similarly varying between 2 or 3 depending on the task, zero weight decay specifically applied to Hyena layers, embedding dropout of 0.2, and residual dropout of 0.2. All three models—Hyena, Transformer++, and Lyra—were trained for 50 epochs per dataset, optimized using AdamW with a learning rate of 0.001 and weight decay of 0.01, guided by cross-entropy loss. Evaluation was conducted using the top-1% accuracy metric for each dataset.

*Epistatic Order Comparison*

We evaluated model performance using a dataset of GFP [32,41], where each protein sequence had an associated binary label per amino acid position, indicating the presence ('1') or absence ('0') of a mutation. After filtering out invalid sequences, we classified each remaining sequence by epistatic order (the number of mutations present) for subsequent analysis. The models were trained to predict fluorescence levels given a GFP sequence. The Lyra model, with 14,417 parameters, was configured with an internal dimension of 48, one PGC layer, one S4D layer, internal dropout of 0.2, and final dropout of 0.1. The Transformer-based baseline, optimized via hyperparameter sweep, had 18,882 parameters and consisted of a single TransformerEncoder layer (model dimension 64, 2 attention heads, dropout=0.1) with RoPE positional embeddings consistent with Transformer++. After training, we analyzed each model's predictive performance as a function of epistatic order to determine their effectiveness in capturing higher-order mutation interactions.

*GFP Transformer vs Lyra Fitness Landscape Visualization*

We visualized protein fitness landscapes derived from a GFP dataset [42] with the goal of predicting fluorescence levels directly from protein sequences. Predictions from Lyra and Transformer-based models were compared by visualizing each model's predicted fluorescence distribution against the true distribution. The Lyra model, consisting of 14,417 parameters, had an internal dimension of 48, one 16-dimensional PGC layer, one S4D layer, internal dropout of 0.2, and final dropout of 0.1. The Transformer model, containing 18,882 parameters, included a single TransformerEncoder layer with two attention heads, a model dimension of 64, positional encoding, and dropout of 0.1; this Transformer configuration was selected via hyperparameter sweep. Lyra more effectively captured the bimodal distribution of fluorescence scores compared to the Transformer.

### *Synthetic Tasks*

*Selective Copying*
The Selective Copying task, a variant of the Copying task designed to evaluate a model's content-aware reasoning capability [31] and was adapted to assess the ability of Lyra, PGC, and Hyena models to detect biologically meaningful signals embedded within longer sequences. We constructed synthetic sequences based on a consensus (wild-type) green fluorescent protein (GFP) sequence, into which biologically-informed mutations were introduced at irregular intervals. Specifically, between 1 and 14 mutations, identified through frequently mutated positions and observed co-mutation patterns in GFP datasets, were embedded sparsely within sequences; remaining positions were filled with zero tokens representing non-informative background regions. Mutations introduced into sequences had Hamming distances ranging from 1 to 28 relative to the wild-type sequence, reflecting biologically realistic diversity. Performance was evaluated across sequence lengths ranging from 64 to 1024 tokens.

Lyra, Hyena, and PGC-only models were tested using a hidden state dimension of 64 and trained for 400,000 steps at each sequence length. Hyperparameter tuning was conducted for Hyena to ensure optimal configuration for comparison. Model performance was assessed based on accuracy in correctly identifying both the positions and identities of embedded mutations.

*Time and Memory Assessments*

We benchmark computational and memory efficiency against pretrained and baseline models—Evo, ProtT5, ESM1B, DistilProtBERT, Hyena, and Lyra—across sequence lengths ranging from 256 to 1,048,576 tokens and batch sizes of 1, 2, and 8 on an Nvidia A100 GPU. All models utilize hardware-optimized kernels: Lyra and Hyena employ FlashFFTConv for efficient depthwise and FFT-based convolutions, while Transformer-based models use Flash Attention for optimized attention computations.

*Synthetic Polynomial Modeling*

We generated synthetic data by sampling inputs uniformly between -1 and 1, then applying a randomly parameterized fifth-degree polynomial with additional sine and cosine terms. To evaluate how effectively each model could learn this function, we used a Transformer and a Lyra architecture, containing 230 and 201 parameters, respectively. The Transformer configuration—consisting of a linear embedding layer (input dimension=1, model dimension=6), positional encoding, and a single-layer Transformer encoder—was determined via hyperparameter sweep, resulting in one attention head, a feed-forward dimension of 1, and no dropout. The Lyra model was configured equivalently, with a model dimension of 4. Both models were trained for 3000 epochs.

*Frequency Analysis*

We introduce an extended synthetic frequency-analysis task, in which models are trained to predict complete composite signals rather than solely regressing their magnitudes [121]. Synthetic sequences were constructed as composite signals by summing four distinct cosine waves, each characterized by unique frequencies, amplitudes (coefficients), and phase offsets. The aim of this task was to evaluate the ability of models to learn and disambiguate sequences composed of multiple overlapping functional signals, each corresponding to distinct underlying frequency components. The Lyra model consisted of an internal dimension of 48, one 16-dimensional PGC layer, one S4D layer, internal dropout of 0.2, and final dropout of 0.1, totaling 14,417 parameters. The Hyena model, whose final configuration was determined via hyperparameter sweep, contained 2 Hyena operator layers, each with a hidden dimension of 64, filter order of 128, and dropout of 0.1, totaling 36,096 parameters. After training, we evaluated model performance by measuring the accuracy of reconstructing the overall composite signals and their effectiveness in individually predicting and separating each embedded frequency component.

**ACKNOWLEDGEMENTS**

This work was made possible by funding provided by the Howard Hughes Medical Institute, and a group of generous donors through TED's Audacious Project, a collaborative funding initiative including The ELMA Foundation, MacKenzie Scott, the Skoll Foundation, and Open Philanthropy. We also gratefully acknowledge support from the Centers for Disease Control and Prevention (no. 75D30124C20246), specifically for the application of Lyra to pathogen diagnostic datasets in this work.

**AUTHOR CONTRIBUTIONS**

Conceptualization: K.R., S.M.S.; Methodology: K.R., S.M.S., M.D.M., P.C.S.; Validation: K.R., S.M.S.; Investigation:K.R., S.M.S., A.G., M.D.M., P.C.S.; Data Curation: K.R., S.M.S.; Writing - Original Draft: K.R., S.M.S.; Writing - Review & Editing: K.R., S.M.S., A.G., M.D.M., P.C.S.; Visualization: K.R., S.M.S.; Supervision: A.G., M.D.M., P.C.S.

**COMPETING INTEREST STATEMENT**

P.C.S. is a co-founder and shareholder of Delve Bio. She was a co-founder and shareholder of Sherlock Biosciences (sold to Orasure in December 2024) and was a non-executive board member and shareholder of Danaher Corporation (stepping down in December 2024). A.G. is a co-founder, shareholder, and Chief Scientist of Cartesia AI. K.R. is a shareholder, and Chief Scientist at Lexmata AI.  S.M.S. and P.C.S. are inventors on patents related to Cas13 diagnostics. K.R., S.M.S., M.D.M., P.C.S. are inventors in the U.S. Provisional Patent Application No. 63/657,738 (Broad Ref: BI-11252, JMIN Ref: BROD-6040P) filed for this work.

**Supplementary Figure**

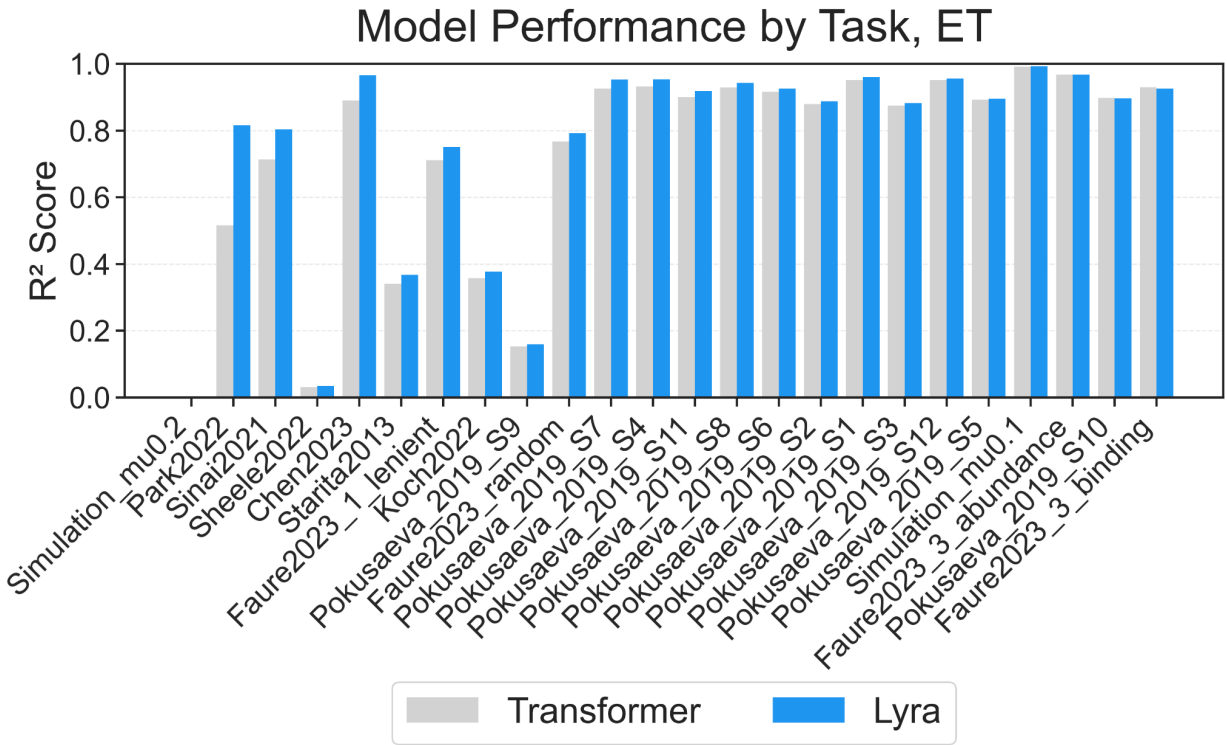

**Supplementary Figure 1**: Performance comparison of Lyra and Transformer models across Epistatic Transformer (ET) tasks, measured by $R^2$ score. Lyra consistently outperforms Transformers across all evaluated tasks, demonstrating its ability to model higher-order dependencies in biological sequences.

# APPENDIX: POLYNOMIAL EXPRESSIVITY

## 1 FORMALIZING BIOLOGICAL SEQUENCE MODELING

Biological macromolecules can be viewed as discrete sequences over a finite alphabet. In the case of proteins, this alphabet typically consists of 20 symbols, each representing an amino acid. Concretely, given a protein sequence, can we predict how that sequence "scores" on a specific trait? Formally, let $\Sigma$ be an alphabet with $|\Sigma| = 20$. We consider sequences of length $l$:

$$u = (u_1, u_2, \ldots, u_l) \quad \text{where} \quad u_i \in \Sigma \text{ for } i = 1, 2, \ldots, l.$$

We then define a real-valued function

$$f : \Sigma^l \to \mathbb{R},$$

which, for a given protein sequence $x$, returns some measured or predicted property such as fluorescence, solubility, or fitness. The essential problem is to learn or approximate $f$ efficiently from a set of experimentally measured sequence–property pairs, despite the potentially large combinatorial space of possible amino acid configurations. A further practical challenge in this learning task is that biological datasets are often comparatively small, limiting the applicability of standard large-scale language-modeling approaches. Instead, more specialized, data-efficient methods are needed to capture the relevant sequence–function relationships when only a limited number of labeled variants are available.

A central complication in modeling these sequence–to–function relationships arises from the fact that real proteins often exhibit dependencies that cannot be captured by simple additive models. Specifically, a mutation at one position may have a different effect depending on the amino acids present at other positions. This phenomenon, known as epistasis, implies that the contribution of a particular mutation depends on context Poelwijk et al. (2019); Zhou & McCandlish (2020); Aghazadeh et al. (2021); Faure et al. (2024), which significantly increases the complexity of learning $f$. An effective way to formalize and analyze such context-dependent effects is to represent $f$ as a multivariate polynomial in encoded variables $\mathbf{u} = (u_1, u_2, \ldots, u_l)$, where each $u_i$ encodes the state of residue $i$. We then write

$$f(\mathbf{u}) = \sum_{k=1}^{K} \sum_{1 \leq i_1 < i_2 < \cdots < i_k \leq l} c_{i_1 i_2 \cdots i_k} \, u_{i_1} \, u_{i_2} \, \cdots \, u_{i_k},$$

with coefficients $c_{i_1 i_2 \cdots i_k} \in \mathbb{R}$ capturing how each subset of residues jointly influences the functional property. When $k = 1$, the polynomial accounts for single-position effects; when $k = 2$, pairwise epistasis enters; and so on. Higher-order terms appear whenever three or more positions collectively modulate function in a way that cannot be reduced to lower-order combinations. From the biological perspective, these polynomial expansions provide a powerful lens for characterizing the underlying fitness landscape—be it protein fluoresence, RNA structure, or CRISPR enzymatic activity—by explicitly capturing the combinatorial interplay among residues.

Because the number of possible interaction terms can grow rapidly with sequence length, an essential aspect of this framework is discovering and approximating these polynomial coefficients in a manner that balances expressiveness with data efficiency. Enumerating all possible subsets of positions becomes infeasible for larger proteins, yet overlooking important interactions degrades predictive accuracy. The practical objective is therefore to uncover the subset of polynomial terms that meaningfully contribute to function without requiring exhaustive measurements of every possible variant. Achieving this objective allows us to model highly nonlinear, context-dependent relationships while remaining tractable in data-constrained regimes. In so doing, the polynomial perspective provides a principled way to think about how multiple amino acids coordinate to yield a given functional outcome, illuminating why purely additive approaches often fail to capture the richness of biological epistasis and highlighting the need for models that can handle sparse, high-dimensional data effectively.

By casting protein sequence–function prediction as the problem of inferring the coefficients in a (potentially high-order) polynomial expansion, we obtain a mathematically transparent representation of how individual residues and their interactions contribute to a measured trait. This representation encodes the combinatorial complexity of biological macromolecules, offering insight into which residues are critical for function and how they cooperate or interfere with each other in shaping a protein's properties. Crucially, it also accommodates the reality of limited experimental data by guiding us toward strategies designed to uncover the most relevant epistatic interactions without demanding prohibitively large datasets.

## 2 MODELING BIOLOGICAL SEQUENCES VIA SSMS

State Space Models (SSMs) provide a structured and efficient approach to sequence modeling by parameterizing the dynamics of a hidden state $x_t$ that evolves over time. In the discrete-time formulation, an SSM is defined as:

$$x_{t+1} = Ax_t + Bu_t, \quad y_t = Cx_t + Du_t,$$

where $x_t \in \mathbb{R}^N$ represents the hidden state, $u_t \in \mathbb{R}$ is the input sequence, and $y_t \in \mathbb{R}$ is the output. The matrices $A \in \mathbb{R}^{N \times N}$, $B \in \mathbb{R}^N$, $C \in \mathbb{R}^{1 \times N}$, and scalar $D$ are learnable parameters Gu et al. (2022; 2021). This formulation provides a compact way to model sequences, where the state $x_t$ encapsulates the input history and evolves through the dynamics encoded by $A$, $B$, and $C$.

The S4D model is a specific diagonal variant of state-space models (SSMs), in which the state transition matrix $A$ is constrained to a diagonal form. In this parameterization, each dimension of the hidden state evolves independently, allowing for straightforward and efficient computation. Specifically, the diagonal entries of the complex-valued $A$ matrix directly control key characteristics of the resulting convolutional filters. The real part of each diagonal entry determines the decay rate of the corresponding filter, governing how rapidly past inputs lose influence. The imaginary part, on the other hand, dictates the frequency of oscillation for each filter, effectively setting the temporal scales at which each hidden dimension processes sequence inputs. Thus, the diagonal structure of S4D not only simplifies computational complexity—enabling efficient inference—but also provides a direct mechanistic link between the parameterization of the state matrix and the temporal dynamics modeled by the network.

State-space models (SSMs) can equivalently be characterized through rational transfer functions Parnichkun et al. (2024), providing a frequency-domain interpretation of their behavior. A rational transfer function expresses the relationship between an input and output sequence through the ratio of two polynomials whose coefficients directly reflect the underlying state-space parameters $(A, B, C)$. Critically, this polynomial structure provides insight into the capacity of SSMs to represent complex, high-order polynomial interactions—such as those arising from biological epistasis—within their hidden state. Evaluating these polynomials at specific frequencies corresponds to examining the impulse response characteristics of the model, effectively translating temporal sequence dynamics into polynomial evaluations. Thus, the rational transfer function framework offers a general and principled lens through which the expressive capacity of SSMs can be analyzed, directly linking polynomial complexity captured by the hidden state dimension to the frequency-domain characteristics encoded within the model.

In light of the connection between diagonalized SSMs such as S4D and their rational transfer function representation, it becomes clear how these models naturally accommodate polynomial interactions among residues within their hidden state. As discussed previously, epistatic interactions among residues can be effectively expressed through polynomial expansions, where the complexity and degree of these polynomials correspond to biologically meaningful combinatorial interactions. Evaluating these polynomial terms explicitly typically becomes computationally prohibitive as interaction complexity grows. The S4D model addresses this challenge by compactly encoding polynomial interactions in a truncated generating function, allowing for efficient evaluation through the discrete Fourier transform.

Formally, the impulse response of an S4D model is given by:

$$h_t = \begin{cases} h_0 & t = 0 \\ CA^{t-1}B & 0 < t \leq \ell \\ 0 & t > \ell, \end{cases}$$

and it can be succinctly expressed as a truncated generating function in the $\mathcal{Z}$-domain:

$$H_\ell(z) = \sum_{t=0}^{\ell-1} h_t z^{-t}.$$

Mechanistically, the hidden state parameters $A, B, C$ encode polynomial terms whose maximal degree directly depends on the dimensionality of the hidden state. By controlling the size of this hidden state, the model explicitly sets a constraint on the complexity of epistatic interactions that it can capture.

In practice, S4D evaluates these polynomial terms efficiently through a frequency-domain computation. Specifically, the truncated generating function $H_\ell(z)$ is evaluated at the complex $m$-th roots of unity, $\mathbb{T}_m = \{e^{2\pi i k/m}\}_{k=0}^{m-1}$. Evaluating polynomials at these specific points is mathematically equivalent to performing a Fast Fourier Transform (FFT) operation on their coefficient vectors. Thus, given polynomial coefficients encoded by the impulse response vector $h = (h_0, h_1, \ldots, h_{\ell-1})$, we have:

$$(H_\ell(z))_{z \in \mathbb{T}_m} = \mathrm{FFT}_m(h).$$

Conversely, the impulse response terms $h_t$ can be recovered precisely through the inverse FFT (iFFT):

$$h_t = \mathrm{iFFT}_m\big((H_\ell(z))_{z \in \mathbb{T}_m}\big)_t.$$

Through this procedure, S4D directly and efficiently evaluates the polynomial terms embedded within its hidden state. The dimension of this state explicitly governs the degree of polynomial (epistatic) interactions that the model is able to represent, providing a clear link between internal model parameters, computational efficiency, and biological interpretability.

## 3 PROJECTED GATED CONVOLUTIONS

Having established how S4D handles long-range dependencies through fast state space kernels, we now introduce a complementary mechanism for local gating: *depthwise 1D convolutions* combined with an *element-wise* (*Hadamard*) product Arora et al. (2023). While S4D specializes in capturing global structure, many architectures rely on gating to modulate signal flow at a more granular level. Specifically, before the S4D layers, we can place a local convolutional block that *gates* or *filters* features through a simple multiplication in each channel.

Depthwise convolution operates on an input $\mathbf{u} \in \mathbb{R}^{l \times d}$, where $l$ denotes the sequence length and $d$ is the number of channels. For a kernel size $k = 3$ (with padding 1), the depthwise convolution applies a separate filter to each channel $c$. Denoting the convolution weights by $\mathbf{W}_{\mathrm{conv}} \in \mathbb{R}^{k \times d}$, the output at position $i$ and channel $c$ is given by

$$\mathbf{u}_{\mathrm{conv}}[i, c] = \sum_{j=-1}^{1} \mathbf{W}_{\mathrm{conv}}[j+1, c]\,\mathbf{u}[i+j, c].$$

Because each channel $c$ has its own slice of $\mathbf{W}_{\mathrm{conv}}$, this operation captures local patterns (*e.g.*, 3-mer motifs in proteins) in a channel-specific manner, with no mixing between different channels at this stage. As a result, depthwise convolution focuses on extracting local features in each dimension while remaining parameter-efficient compared to a full convolution.

To incorporate global channel interactions at each position, we apply a position-wise linear layer $\mathbf{W}_{\mathrm{lin}} \in \mathbb{R}^{d \times d}$ and bias $\mathbf{b}_{\mathrm{lin}} \in \mathbb{R}^d$. For each position $i$ and channel $c$,

$$\mathbf{u}_{\mathrm{lin}}[i, c] = \sum_{c'=1}^{d} \mathbf{W}_{\mathrm{lin}}[c, c']\,\mathbf{u}[i, c'] \;+\; \mathbf{b}_{\mathrm{lin}}[c].$$

Unlike the convolution step, which aggregates nearby positions within the same channel, this linear transformation does not look at neighboring indices $i \pm 1$; instead, it mixes features across all channels $c'$ at the same position $i$. In a biological context, this step can learn how different channel encodings (*e.g.*, properties of amino acids, or hidden representation dimensions) should be combined or reweighted based on the broader embedding at that position.

Combining these two outputs through a Hadamard product (elementwise multiplication) yields a gating mechanism. Concretely, the final output is defined by

$$\mathbf{u}_{\text{out}}[i,c] = \mathbf{u}_{\text{conv}}[i,c] \cdot \mathbf{u}_{\text{lin}}[i,c].$$

Substituting the definitions of $\mathbf{u}_{\text{conv}}$ and $\mathbf{u}_{\text{lin}}$ into this product exposes how second-order interactions arise. Specifically,

$$\mathbf{u}_{\text{out}}[i,c] = \left( \sum_{j=-1}^{1} \mathbf{W}_{\text{conv}}[j+1, c]\, \mathbf{u}[i+j, c] \right) \times \left( \sum_{c'=1}^{d} \mathbf{W}_{\text{lin}}[c, c']\, \mathbf{u}[i, c'] + \mathbf{b}_{\text{lin}}[c] \right).$$

Expanding this product shows that local features $\mathbf{u}[i+j,c]$ (extracted by depthwise convolution) multiply with channel-mixed features $\mathbf{u}[i,c']$ (from the linear layer) at the same position $i$. If we look more closely at the cross-terms,

$$\sum_{j=-1}^{1} \sum_{c'=1}^{d} \mathbf{W}_{\text{conv}}[j+1, c]\, \mathbf{W}_{\text{lin}}[c, c']\, \mathbf{u}[i+j, c]\, \mathbf{u}[i, c'],$$

we see an explicit second-order interaction between $\mathbf{u}[i+j, c]$ and $\mathbf{u}[i, c']$. In epistatic terms, a residue (or embedding feature) at position $i+j$ interacts multiplicatively with features at the same position $i$ but different channel $c'$. This captures context-dependent local patterns: whether a motif at position $i+j$ influences the output depends on how the channels at position $i$ are activated, and vice versa. Stacking multiple such layers in sequence enables even richer, higher-order interactions because each subsequent layer's input can include products of features formed at earlier layers. Hence, by combining a depthwise convolution for local context, a linear transformation for channel mixing, and a Hadamard product for gating, we effectively construct a module that models second-order epistatic effects in a compact, parameter-efficient manner and allows for further compositional complexity in deeper architectures.

# 4 CODE IMPLEMENTATIONS

## 4.1 PROJECTED GATED CONVOLUTION:

```python
import torch
import torch.nn as nn

class PGC(nn.Module):
    def __init__(self,d_model,expansion_factor = 1.0,dropout = 0.0):
        super().__init__()
        self.d_model = d_model
        self.expansion_factor = expansion_factor
        self.dropout = dropout
        self.conv = nn.Conv1d(d_model, d_model, kernel_size=3, padding=1, groups=d_model)
        self.in_proj = nn.Linear(d_model, int(d_model * expansion_factor *2))
        self.norm = nn.RMSNorm(int(d_model * expansion_factor))
        self.in_norm = nn.RMSNorm(d_model * expansion_factor * 2)
        self.out_proj = nn.Linear(int(d_model * expansion_factor), d_model)
        self.dropout = nn.Dropout(dropout)
    def forward(self, u):
        xv = self.in_norm(self.in_proj(u))
        x,v = xv.chunk(2,dim=-1)
        x_conv = self.conv(x.transpose(-1,-2)).transpose(-1,-2)
        gate =  v * x_conv
        x = self.norm(self.out_proj(gate))
        return x
```

Listing 1: PGC: Project Gated Convolution Class

## 4.2 S4D:

```python
import math
import torch
import torch.nn as nn
import torch.nn.functional as F
from einops import rearrange, repeat

class DropoutNd(nn.Module):
    def __init__(self, p: float = 0.5, tie=True, transposed=True):
        super().__init__()
        if p < 0 or p >= 1:
            raise ValueError("dropout probability has to be in [0, 1), " "but got {}".format(p))
        self.p = p
        self.tie = tie
        self.transposed = transposed

    def forward(self, X):
        if self.training:
            if not self.transposed: X = rearrange(X, 'b ... d -> b d ...')
            mask_shape = X.shape[:2] + (1,)*(X.ndim-2) if self.tie else X.shape
            mask = torch.rand(*mask_shape, device=X.device) < 1.-self.p
            X = X * mask * (1.0/(1-self.p))
            if not self.transposed: X = rearrange(X, 'b d ... -> b ... d')
            return X

class S4DKernel(nn.Module):
    def __init__(self, d_model, N=64, dt_min=0.001, dt_max=0.1, lr=None):
```

```python
27          super().__init__()
28          H = d_model
29          log_dt = torch.rand(H) * (math.log(dt_max) - math.log(dt_min)) +
    math.log(dt_min)
30          C = torch.randn(H, N // 2, dtype=torch.cfloat)
31          self.C = nn.Parameter(torch.view_as_real(C))
32          self.register("log_dt", log_dt, lr)
33          log_A_real = torch.log(0.5 * torch.ones(H, N//2))
34          A_imag = math.pi * repeat(torch.arange(N//2), 'n -> h n', h=H)
35          self.register("log_A_real", log_A_real, lr)
36          self.register("A_imag", A_imag, lr)
37
38     def forward(self, L):
39          dt = torch.exp(self.log_dt)
40          C = torch.view_as_complex(self.C)
41          A = -torch.exp(self.log_A_real) + 1j * self.A_imag
42          dtA = A * dt.unsqueeze(-1)
43          K = dtA.unsqueeze(-1) * torch.arange(L, device=A.device)
44          C = C * (torch.exp(dtA)-1.) / A
45          K = 2 * torch.einsum('hn, hnl -> hl', C, torch.exp(K)).real
46          return K
47
48     def register(self, name, tensor, lr=None):
49          if lr == 0.0:
50              self.register_buffer(name, tensor)
51          else:
52              self.register_parameter(name, nn.Parameter(tensor))
53              optim = {"weight_decay": 0.0}
54              if lr is not None: optim["lr"] = lr
55              setattr(getattr(self, name), "_optim", optim)
56
57 class S4D(nn.Module):
58     def __init__(self, d_model, d_state=64, dropout=0.0, transposed=True,
     **kernel_args):
59          super().__init__()
60          self.h = d_model
61          self.n = d_state
62          self.d_output = self.h
63          self.transposed = transposed
64          self.D = nn.Parameter(torch.randn(self.h))
65          self.kernel = S4DKernel(self.h, N=self.n, **kernel_args)
66          self.activation = nn.GELU()
67          self.dropout = DropoutNd(dropout) if dropout > 0.0 else nn.
     Identity()
68          self.output_linear = nn.Sequential(
69              nn.Conv1d(self.h, 2*self.h, kernel_size=1),
70              nn.GLU(dim=-2),
71          )
72
73     def forward(self, u, **kwargs):
74          if not self.transposed: u = u.transpose(-1, -2)
75          L = u.size(-1)
76          k = self.kernel(L=L)
77          k_f = torch.fft.rfft(k, n=2*L)
78          u_f = torch.fft.rfft(u, n=2*L)
79          y = torch.fft.irfft(u_f*k_f, n=2*L)[..., :L]
80          y = y + u * self.D.unsqueeze(-1)
81          y = self.dropout(self.activation(y))
82          y = self.output_linear(y)
83          if not self.transposed: y = y.transpose(-1, -2)
84          return y
```

Listing 2: Minimal S4D Implementation for Pedagogical Purposes

### 4.3 LYRA FULL IMPLEMENTATION:

```
class Lyra(nn.Module):
    """
    Lyra model incorporating PGC and S4D layers for sequence processing.

    Parameters:
    - model_dimension: Internal dimension of S4D layers and projection
    layers before and after PGC.
    - pgc_configs (list of tuples): Configuration for PGC layers, where
    each tuple contains
      (hidden dimension, number of layers in the PGC module).
    - num_s4 (int): Number of S4D layers.
    - d_input (int): Dimensionality of the input features.
    - d_output (int, optional): Dimensionality of the output features.
    Defaults to 10.
    - dropout (float, optional): Dropout rate for regularization.
    Defaults to 0.2.
    - prenorm (bool, optional): Whether to use pre-normalization.
    Defaults to True.
    """

    def __init__(
        self,
        model_dimension,
        pgc_configs,
        num_s4,
        d_input,
        d_output=10,
        dropout=0.2,
        prenorm=True,
        final_dropout=0.2
    ):
        super().__init__()
        self.encoder = nn.Linear(d_input, model_dimension)
        self.pgc_layers = nn.ModuleList()
        for config in pgc_configs:
            pgc_hidden_dimension, num_layers = config
            self.pgc_layers.append(PGC(model_dimension, pgc_hidden_dimension,
                                      pgc_hidden_dimension, dropout, num_layers))

        self.prenorm = prenorm

        # Stack S4 layers as residual blocks
        self.s4_layers = nn.ModuleList()
        self.norms = nn.ModuleList()
        self.dropouts = nn.ModuleList()
        for _ in range(num_s4):
            self.s4_layers.append(
                S4D(model_dimension, dropout=dropout, transposed=True, lr
=min(0.001, 0.002))
            )
            self.norms.append(RMSNorm(model_dimension))
            self.dropouts.append(dropout_fn(dropout))

        # Linear decoder
        self.decoder = nn.Linear(model_dimension, d_output)
        self.dropout = nn.Dropout(final_dropout)

    def forward(self, x, return_embeddings=False):
        """
        Input x is shape (B, L, d_input)
        """
```

```python
        x = self.encoder(x)   # (B, L, d_input) -> (B, L, d_model)
        for pgc_layer in self.pgc_layers:
            x = pgc_layer(x)
        x = x.transpose(-1, -2)   # (B, L, d_model) -> (B, d_model, L)
        for layer, norm, dropout in zip(self.s4_layers, self.norms, self.dropouts):
            z = x
            if self.prenorm:
                # Prenorm
                z = norm(z.transpose(-1, -2)).transpose(-1, -2)
            # Apply S4 block
            z = layer(z)
            # Dropout on the output of the S4 block
            z = dropout(z)
            # Residual connection
            x = z + x
            if not self.prenorm:
                # Postnorm
                x = norm(x.transpose(-1, -2)).transpose(-1, -2)

        x = x.transpose(-1, -2)

        embeddings = x

        # Pooling: average pooling over the sequence length
        x = x.mean(dim=1)

        # Decode the outputs
        x = self.dropout(x)   # (B, d_model) -> (B, d_output)
        x = self.decoder(x)

        if return_embeddings:
            return x, embeddings
        else:
            return x
```

Listing 3: Lyra Model Combining PGC and S4D Layers